\documentclass{article}

\usepackage{arxiv}

\usepackage{amsmath}
\usepackage{graphicx}
\usepackage{dcolumn}
\usepackage{bm}
\usepackage{amssymb}

\usepackage[utf8]{inputenc}
\usepackage[T1]{fontenc}
\usepackage{mathptmx}
\usepackage{etoolbox}
\usepackage{caption}
\usepackage{subcaption}

\usepackage[T1]{fontenc}    
\usepackage{hyperref}       
\usepackage{nameref}        
\usepackage{url}            
\usepackage{booktabs}       
\usepackage{amsfonts}       
\usepackage{nicefrac}       
\usepackage{microtype}      
\usepackage{multirow}
\usepackage{fancyhdr}       
\usepackage{graphicx}    
\usepackage{geometry}
\usepackage{xcolor}
\usepackage{algorithm}
\usepackage{algorithmic}

\usepackage{float}

\usepackage{booktabs}
\usepackage{enumitem}
\usepackage{lineno}

\usepackage{nomencl}
\makenomenclature
\usepackage{etoolbox}

\renewcommand\nomgroup[1]{%
  \item[\bfseries
  \ifstrequal{#1}{G}{Geometry}{%
  \ifstrequal{#1}{D}{Dimensional}{%
  \ifstrequal{#1}{N}{Non-dimensional}{%
  \ifstrequal{#1}{P}{PINN related}{%
  \ifstrequal{#1}{A}{Acronyms}{}}}}}%
]}
\usepackage[numbers,sort&compress]{natbib}
\usepackage{doi}
\usepackage{multicol}
\usepackage{authblk}
\usepackage{hyperref}

\title{Parameter Estimation in Machining Dynamics with Regenerative Delay and Nonsmooth Friction using Physics-Informed Neural Networks}

\date{}

\newbox{\orcid}\sbox{\orcid}{\includegraphics[scale=0.06]{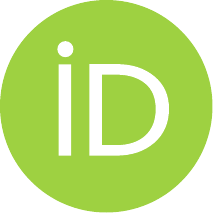}}

\author[1]{%
	\href{https://orcid.org/0000-0001-8971-965X}{\usebox{\orcid}\hspace{1mm}Meiyazhagan Jaganathan}%
}
\author[1]{%
    \href{https://orcid.org/0000-0002-8318-3521}
    {\usebox{\orcid}\hspace{1mm}Vikram Pakrashi}%
}

\author[1]{%
	\href{https://orcid.org/0000-0003-1497-145X}{\usebox{\orcid}\hspace{1mm}Aasifa Rounak\thanks{corresponding author - \texttt{aasifa.rounak@ucd.ie}}}%
}

\affil[1]{UCD Centre for Mechanics, Dynamical Systems and Risk Laboratory, School of Mechanical and Materials
Engineering, University College Dublin, D04 V1W8, Dublin, Ireland}

\renewcommand{\headeright}{}
\renewcommand{\undertitle}{}
\renewcommand{\shorttitle}{Parameter Estimation in Machining Dynamics with Regenerative Delay and Nonsmooth Friction using Physics-Informed Neural Networks}

\hypersetup{
pdftitle={Parameter Estimation in Machining Dynamics with Regenerative Delay and Nonsmooth Friction using Physics-Informed Neural Networks},
pdfsubject={Meiyazhagan Jaganathan}
}

\begin{document}
\maketitle

\begin{abstract}
A multi-domain eXtended Physics-Informed Neural Network (XPINN) framework is developed for nonsmooth Delay Differential Equations (DDEs). This is the first implementation to demonstrate the efficacy of partitioning the temporal domain into subdomains of integer multiples of the characteristic time delay and progressively training the associated subnetworks while freezing previously learned parameters. The efficacy of the proposed framework is demonstrated using a machining dynamics model that incorporates both regenerative and nonsmooth frictional effects. Results demonstrate that the proposed multi-domain XPINN framework leads to better solution reconstruction in DDEs and improved parameter estimation compared to a generic PINN (SPINN) formulation. The proposed method works particularly well for extended temporal domains and non-constant history functions. The robustness of inverse XPINN (I-XPINN) is also assessed using reference data contaminated with Gaussian measurement noise. Results indicate that I-XPINN remains resilient to measurement noise and the physics-informed constraints guide the network toward accurately recovering the underlying dynamics. This demonstrates, for the first time, the potential of the proposed framework for reliable parameter identification in DDEs characterised by nonsmoothness and large time delays.
\end{abstract}

\vspace{2em}


\nomenclature[D]{$Y(t)$}{Tool displacement [m]}
\nomenclature[D]{$\dot{Y}(t)$}{Tool velocity [m\,s$^{-1}$]}
\nomenclature[D]{$\ddot{Y}(t)$}{Tool acceleration [m\,s$^{-2}$]}
\nomenclature[D]{$m$}{Tool mass [kg]}
\nomenclature[D]{$c$}{Damping coefficient [N\,s\,m$^{-1}$]}
\nomenclature[D]{$k$}{Stiffness [N\,m$^{-1}$]}
\nomenclature[D]{$D$}{Workpiece diameter [m]}
\nomenclature[D]{$R$}{Workpiece radius, $R = D/2$ [m]}
\nomenclature[D]{$N$}{Spindle speed [rev\,min$^{-1}$]}
\nomenclature[D]{$\mu$}{Friction coefficient [-]}
\nomenclature[D]{$\mu_s$}{Static friction coefficient [-]}
\nomenclature[D]{$\mu_d$}{Dynamic friction coefficient [-]}
\nomenclature[D]{$V_c$}{Cutting speed, $V_c=2\pi RN/60$ [m\,s$^{-1}$]}
\nomenclature[D]{$V_{ch}$}{Chip velocity [m\,s$^{-1}$]}
\nomenclature[D]{$V_\gamma$}{Relative chip velocity, $V_\gamma=V_{ch}-\dot{Y}\cos\gamma$ [m\,s$^{-1}$]}
\nomenclature[D]{$V_s$}{Stribeck velocity [m\,s$^{-1}$]}
\nomenclature[D]{$K$}{Cutting force coefficient [N\,m$^{-2}$]}
\nomenclature[D]{$a_p$}{Depth of cut [m]}
\nomenclature[D]{$C_y$}{Process damping coefficient [N\,s\,m$^{-2}$]}
\nomenclature[D]{$H_D$}{Nominal chip thickness (feed) [m]}
\nomenclature[D]{$H(t)$}{Instantaneous chip thickness [m]}
\nomenclature[D]{$F_\gamma$}{Tangential force on rake face [N]}
\nomenclature[D]{$F_{\gamma n}$}{Normal force on rake face [N]}
\nomenclature[D]{$F_r$}{Resultant force [N]}
\nomenclature[D]{$F_c$}{Cutting force ($x$-direction) [N]}
\nomenclature[D]{$F_f$}{Feed force ($y$-direction) [N]}
\nomenclature[D]{$F_d$}{Process damping force ($y$-direction) [N]}
\nomenclature[D]{$t_w$}{Rotational period of the workpiece $t_w={60}/{N}$ [s] (time-delay)}


\nomenclature[N]{$\tau$}{Non-dimensional time, $\tau=t\sqrt{k/m}$ [-]}
\nomenclature[N]{$\tau_w$}{Non-dimensional delay, $\tau_w=60/n$ [-]}
\nomenclature[N]{$y(\tau)$}{Non-dimensional displacement, $y=Y/H_D$ [-]}

\nomenclature[N]{$y_1(\tau)$}{State 1, $y_1=Y/H_D$ [-]}
\nomenclature[N]{$y_2(\tau)$}{State 2, $y_2=\dot{y}_1$ [-]}
\nomenclature[N]{$\dot{y}(\tau)$}{Non-dimensional velocity [-]}
\nomenclature[N]{$n$}{Non-dimensional spindle speed, $n = N\sqrt{m/k}$ [-]}
\nomenclature[N]{$c_y$}{Non-dimensional process damping, $c_y=30C_y/(\pi RK)$ [-]}
\nomenclature[N]{$\mu_{vs}$}{Non-dimensional friction coefficient [-]}
\nomenclature[N]{$v_s$}{Non-dimensional Stribeck velocity [-]}
\nomenclature[N]{$\nu$}{Non-dimensional chip thickness parameter [-]}
\nomenclature[N]{$v_\gamma$}{Non-dimensional relative velocity [-]}
\nomenclature[N]{$\xi$}{Damping ratio, $\xi = c/\sqrt{mk}$ [-]}
\nomenclature[N]{$g$}{Friction law argument [-]}
\nomenclature[N]{$\Omega$}{Non-dimensional frequency, $\Omega = 2\pi/\tau_w$ [-]}

\nomenclature[N]{$W$}{Non-dimensional depth parameter, $W = a_pK/k$ [-]}
\nomenclature[N]{$h(\tau)$}{Non-dimensional chip thickness, $h=1-y+y(\tau-\tau_w)$ [-]}
\nomenclature[N]{$f_f$}{Non-dimensional cutting force [-]}
\nomenclature[N]{$f_d$}{Non-dimensional damping force [-]}

\nomenclature[G]{$\gamma$}{Rake angle [degrees]}
\nomenclature[G]{$\phi$}{Shear angle [degrees]}


\nomenclature[A]{SDOF}{Single Degree of Freedom}
\nomenclature[A]{DDE}{Delay Differential Equation}
\nomenclature[A]{AD}{Automatic Differentiation}
\nomenclature[A]{ANN}{Artificial Neural Network}
\nomenclature[A]{PINN}{Physics-Informed Neural Network}

\nomenclature[A]{SPINN}{Single-domain Physics-Informed Neural Network}
\nomenclature[A]{I-SPINN}{Inverse Single-domain Physics-Informed Neural Network}
\nomenclature[A]{XPINN}{eXtended Physics-Informed Neural Network}
\nomenclature[A]{I-XPINN}{Inverse eXtended Physics-Informed Neural Network}
\nomenclature[A]{SLD}{{Stability Lobe Diagram}}
\nomenclature[A]{RPM}{{Revolutions Per Minute}}

\nomenclature[P]{$n_L$}{Number of hidden layers}
\nomenclature[P]{$y_\theta(\tau)$}{ANN approximated solution at $\tau$}
\nomenclature[P]{$\mathcal{N}_\theta$}{ANN approximation function}
\nomenclature[P]{$\mathcal{N}_\theta^{(k_d)}$}{ANN subnet for sub-domain $k_d$}
\nomenclature[P]{$y_{\mathrm{ref}}(\tau)$}{Reference solution from MATLAB DDE23 solver}
\nomenclature[P]{$y_{\mathrm{ref}}^{\mathrm{noise}}(\tau)$}{Noisy reference solution with measurement uncertainty}
\nomenclature[P]{$y_{\mathrm{pre}}(\tau)$}{PINN prediction of the solution}
\nomenclature[P]{$\theta$}{Trainable parameters of the ANN}

\nomenclature[P]{$\tau_{\mathrm{start}}$}{Start of physical time interval}
\nomenclature[P]{$\tau_{\mathrm{end}}$}{End of physical time interval}
\nomenclature[P]{$\dot{y}_\theta(\tau)$}{First derivative of $y_\theta$ with respect to $\tau$}
\nomenclature[P]{$\ddot{y}_\theta(\tau)$}{Second derivative of $y_\theta$ with respect to $\tau$}
\nomenclature[P]{$Z_l$}{Output of the $l^{\mathrm{th}}$ hidden layer}
\nomenclature[P]{$w_l$}{Weight matrix of the $l^{\mathrm{th}}$ layer}
\nomenclature[P]{$b_l$}{Bias vector of the $l^{\mathrm{th}}$ layer}
\nomenclature[P]{$d_l$}{Dimension of $Z_l$, i.e., $\dim(Z_l)$}
\nomenclature[P]{$n_L$}{Number of hidden layers}
\nomenclature[P]{$\mathbf{Z}$}{Hidden state vector}
\nomenclature[P]{$\mathbf{Z}_r$}{Residual branch output in a residual block}
\nomenclature[P]{$\mathbf{Z}_{\text{out}}$}{Output of a residual block}
\nomenclature[P]{$\mathbf{W}_1, \mathbf{W}_2$}{Trainable weight matrices in a residual block}
\nomenclature[P]{$\mathbf{b}_1, \mathbf{b}_2$}{Trainable bias vectors in a residual block}

\nomenclature[P]{$K_{\log}$}{Logarithm of $K$, i.e., $\log(K)$}
\nomenclature[P]{$C_{y,\log}$}{Logarithm of $C_y$, i.e., $\log(C_y)$}
\nomenclature[P]{$K^{\mathrm{train}}_{\log}$}{Trainable logarithmic cutting force coefficient}
\nomenclature[P]{$C^{\mathrm{train}}_{y,\log}$}{Trainable logarithmic process damping coefficient}
\nomenclature[P]{$K_{\mathrm{pred}}$}{Predicted cutting force coefficient from PINN, $K_{\mathrm{pred}} = \exp(K_{\log})$}
\nomenclature[P]{$C_{y, \mathrm{pred}}$}{Predicted process damping coefficient from PINN, $C_{y, \mathrm{pred}} = \exp(C_{y, \log})$}
\nomenclature[P]{$K_{\mathrm{true}}$}{True reference value of $K$}
\nomenclature[P]{$C_{y,\mathrm{true}}$}{True reference value of $C_y$}
\nomenclature[P]{$E_K$}{Relative percentage error in $K$}
\nomenclature[P]{$E_{C_y}$}{Relative percentage error in $C_y$}
\nomenclature[P]{$\mathcal{L}_{\mathrm{Total}}$}{Total loss function}
\nomenclature[P]{$\mathcal{L}_{\mathrm{PHY}}$}{Physical loss from governing DDE}
\nomenclature[P]{$\mathcal{L}_I$}{Initial condition loss function}
\nomenclature[P]{$\mathcal{L}_{\mathrm{SH}}$}{Shifted history anchoring loss}
\nomenclature[P]{$\mathcal{L}_{\mathrm{INT}}$}{Interface loss (continuity between sub-domains)}
\nomenclature[P]{$\mathcal{L}_{\mathrm{DATA}}$}{Data loss term}
\nomenclature[P]{$N_f$}{Number of collocation points}
\nomenclature[P]{$N_H$}{Number of points used in shifted-history anchoring loss $\mathcal{L}_{\mathrm{SH}}$}
\nomenclature[P]{$N_{\mathrm{data}}$}{Number of data points used in data loss $\mathcal{L}_{\mathrm{DATA}}$}
\nomenclature[P]{$N_{\mathrm{cand}}$}{Number of candidate points}
\nomenclature[P]{$\mathcal{R}(\tau_i)$}{Residual of the DDE at collocation point $\tau_i$}

\nomenclature[P]{$\Phi_H(\tau)$}{Prescribed history function for $\tau \leq \tau_w$}
\nomenclature[P]{$\tau_i^{(H)}$}{History points in $[0,\tau_w]$}
\nomenclature[P]{$W_P$}{Adaptive weight for physical loss in SPINN}
\nomenclature[P]{$W_I$}{Adaptive weight for initial condition loss in SPINN}
\nomenclature[P]{$W_D$}{Adaptive weight for data loss in SPINN}
\nomenclature[P]{$\lambda_{\mathrm{PHY}}$}{Adaptive weight for physical loss in XPINN}
\nomenclature[P]{$\lambda_{\mathrm{SH}}$}{Adaptive weight for shifted history loss in XPINN}
\nomenclature[P]{$\lambda_{\mathrm{INT}}$}{Adaptive weight for interface loss in XPINN}
\nomenclature[P]{$\lambda_{D}$}{Adaptive weight for data loss in XPINN}
\nomenclature[P]{$\Gamma$}{Small positive regularization constant}
\nomenclature[P]{$\operatorname{SG}(\cdot)$}{Stop-gradient operator}
\nomenclature[P]{$L_2$}{Relative $L_2$ error between prediction and reference}
\nomenclature[P]{$\mathrm{MAE}$}{Mean Absolute Error over collocation points}
\nomenclature[P]{$k_d$}{Sub-domain index}
\nomenclature[P]{$s(\tau)$}{Normalized time variable in $[-1,1]$}
\nomenclature[P]{$\tau(s)$}{Inverse mapping from normalized to physical time}
\nomenclature[P]{$\mathcal{F}(s)$}{Fourier feature embedding of normalized input $s$}
\nomenclature[P]{$\mathbf{B}_{k_d}$}{Fixed random projection matrix for Fourier features}
\nomenclature[P]{$\eta_{epo}$}{Learning rate at epoch ($epo$) used in XPINN}
\nomenclature[P]{$\eta_{0}$}{Initial learning rate used in XPINN}
\nomenclature[P]{$\gamma_d$}{Learning-rate decay factor ($\gamma_d < 1$) for XPINN}
\nomenclature[P]{$\gamma_{\mathrm{lr}}$}{Exponential learning-rate decay factor (scheduler) for SPINN}
\nomenclature[P]{$\sigma_n$}{Noise standard deviation scale}
\nomenclature[P]{$\alpha_n$}{Prescribed noise intensity (fraction of signal std)}
\nomenclature[P]{$\operatorname{\mathbf{std}}(y_{\mathrm{ref}})$}{Standard deviation of the reference solution}
\nomenclature[P]{$\epsilon_n(\tau)$}{Standard Gaussian noise, $\epsilon_n \sim \mathcal{N}(0,1)$}

\printnomenclature[5cm]

\section{Introduction}

\noindent Chatter remains one of the critical challenges in machining. Self-excited vibrations generated during cutting can severely compromise surface finish, reduce dimensional accuracy, accelerate tool wear, and limit productivity \cite{review2024zhang}. This reflects the inherently dynamic nature of machining, where the cutting tool and workpiece interact through forces that depend on both the instantaneous and past states of the system. The resulting dynamics is thus generated from complex nonlinear interactions between the structural response of the machine-tool system and the cutting process. In machining operations, turning is particularly well suited for such investigations because of its relatively simple geometry, well-defined cutting kinematics, and pronounced regenerative mechanism. This configuration enables controlled experiments, reliable force measurements, and direct comparison with mathematical models, making it a representative platform for analysing the chatter vibrations.
 
\par A characteristic feature of machining chatter is the regenerative effect, where the instantaneous chip thickness is influenced by the surface waviness profile left from the previous pass, introducing intrinsic time delay into the governing dynamics. The cutting force depends on both the current and the past tool positions, which naturally leads to the need for delay differential equation (DDE) models \cite{turning-wahi2008self, turning-stepan2001modelling}. Friction between the workpiece and the tool's rake face introduces additional nonlinearities in the cutting forces \cite{turning-wiercigroch2001frictional, turning-rusinek2014modelling}. The combined influence of regenerative and frictional effects results in a nonsmooth dynamical system with memory, whose mathematical modelling is useful not only for predicting stable and unstable cutting responses but also for investigating the mechanisms underlying transitions between different dynamical states \cite{jaganathan2026interplay, turning-yan2021safety}.

\par While DDE models provide a systematic framework for investigating stability, bifurcations, and the onset of chatter for turning by explicitly incorporating delayed cutting-force characteristics, the predictive capability of these models depends critically on the accurate estimation of their physical parameters. In particular, parameters such as the cutting force coefficient and the process damping coefficient determine the magnitude of the cutting forces and the strength of the associated machine-tool interactions, directly impacting the system response and stability predictions. Accurate estimation of these parameters is therefore important for establishing a reliable mathematical model. However, identifying such parameters is challenging because they depend on several factors, such as cutting conditions, tool geometry, cutting speed, tool wear, and flank-face interaction. Conventional identification approaches commonly rely on extensive experimentation-based calibrations or optimisation-based fitting requiring repeated model evaluations \cite{altintas2008identification, feng2023mechanism}. Furthermore, measurement noise, limited observability, nonlinear force characteristics, and the sensitivity of delayed dynamics can severely hinder accurate parameter estimation. Consequently, identifying the underlying machining system parameters from measured system responses constitutes a challenging inverse problem. This difficulty motivates the development of computational approaches that can exploit both the available system-response data and the known underlying physics governing the evolution of the dynamical system.

\par Physics‑Informed Neural Networks (PINNs) provide a powerful framework for tackling such inverse problems. Introduced by Raissi et al.~\cite{raissi2019physics} in 2019, PINNs have since gained widespread adoption due to their ability to incorporate physical knowledge directly into the learning process across numerous scientific and engineering domains, including structural mechanics \cite{thawon2026physics, mai2024physics, zhou2026data}, bio-engineering \cite{kissas2020machine, sahli2020physics, herrero2022ep}, environmental engineering \cite{di2026physics}, fluid mechanics \cite{cai2021physics, raissi2019deepvortex}, nonlinear wave equations \cite{jaganathan2023data, meiyazhagan2022data}, stochastic differential equations \cite{yang2020physics, shin2023physics}, biomechanics \cite{ferron2025physics}, disease spreading dynamics \cite{malinzi2022determining, kharazmi2021identifiability}, and micromechanics \cite{xiong2024physics}. The application of PINNs to machining problems has also attracted increasing attention. For example, tool wear monitoring \cite{zhu2023physics}, prediction of fracture surface quality \cite{liu2025study}, milling vibration prediction \cite{zeng2026physics}, and crater formation and wear-mechanism prediction in Ti6Al4V machining\cite{liu2026physics}. PINN-based frameworks have also been developed for adhesive tool-wear prediction in high-speed milling, with particular emphasis on improving robustness against inaccurate physical models \cite{tan2026physics}, as well as for tool-wear and remaining useful life prediction in deep-hole boring through adaptive loss weighting \cite{zhang2026physics}. More recently, PINN approaches have been applied to five-axis machining for surface partitioning and tool orientation optimisation \cite{hao2025neural}, milling power-consumption prediction \cite{pratap2026development}, and milling surface-roughness prediction with uncertainty quantification \cite{yuan2026precisionpinn}. These studies demonstrate the growing potential of PINN-based frameworks across various applications in machining processes. However, existing applications have predominantly focused on forward prediction and process optimisation of machining systems, whereas parameter estimation of a machining system governed by delayed nonlinear dynamics remains unexplored.

\par The challenges with parameter estimation become more pronounced when PINNs are applied to DDEs. Unlike ordinary differential equations, in which the system state at the current time determines the instantaneous evolution, a DDE depends explicitly on the solution over a preceding time interval. Consequently, approximation using a given physics information must stay consistent with both the current state and the required history of the solution. Recent studies have explored different strategies for incorporating delay information into PINN formulations. He and Tiamiyu \cite{he2025physics} developed a PINN approach for solving DDEs, using neural-network trial functions to satisfy the initial conditions and an iterative formulation to obtain accurate solutions. Their approach was demonstrated through six numerical examples involving scalar and coupled DDE systems with different nonlinear and delayed formulations. Wang et al. \cite{wang2025deep} subsequently proposed a PINN framework to solve forward DDEs and approximate inverse solutions while identifying unknown delay parameters. These studies establish the overall adequacy of PINN-based approaches for both forward solution approximation and inverse parameter identification in DDEs. However, existing studies remain inadequate for addressing the interplay of large delays, extended temporal domains, non-constant history functions, and nonsmooth effects, which presents a significant unresolved challenge in the analysis of such systems. Furthermore, the robustness of PINNs for inverse parameter identification in the presence of measurement noise remains insufficiently explored. These gaps are particularly relevant to machining as it simultaneously involves regenerative delay, nonlinear cutting forces, nonsmooth frictional effects, and process damping, while the characteristic delay is relatively large compared with the temporal scales considered in previously mentioned PINN-DDE studies. Such large delays introduce additional challenges in representing the history-dependent solution and maintaining accurate propagation of the delayed state over the computational domain. 

\par To address this gap, this work first investigates the capability of a single-domain PINN (SPINN), based on the formulation introduced by Wang et al. \cite{wang2025deep}, for approximating the nonlinear machining DDE. The performance of the single-domain formulation is examined for constant and non-constant history functions and over increasingly extended temporal domains. While SPINN can adequately represent delayed dynamics when a constant history function is prescribed, its performance deteriorates substantially when a non-constant history function is used and over extended time intervals. This limitation is significant for the machining problem under consideration because the history function provides the initial delayed state from which the subsequent regenerative dynamics evolve. Therefore, its accurate representation is essential for reliable prediction of the system response. Motivated by these limitations, a multi-domain eXtended Physics-Informed Neural Network (XPINN) framework is then developed, where the temporal domain is decomposed into multiple subdomains, and the solution is learned through interconnected neural networks. The domain-decomposition strategy enables the solution of the delayed system to be approximated locally across successive temporal intervals while maintaining consistency between neighbouring subdomains, thereby improving the representation of machining dynamics over extended time horizons. The forward solution problem is considered first because reliable parameter estimation requires an accurate representation of the underlying delayed dynamics. The resulting framework is then employed to identify the unknown parameter values. The proposed approach is evaluated for both stable and unstable dynamical regimes, and its robustness is further examined in the presence of noise. To the best of the authors’ knowledge, this is the first XPINN-based parameter identification for a nonlinear delayed machining system incorporating regenerative and nonsmooth frictional effects, while explicitly examining the influence of non-constant history functions, extended temporal domains, dynamical regimes, and observational noise.

\par The remainder of this paper is organised as follows. Section.~\ref{sec2} presents the mathematical formulation of the metal cutting system with regenerative and frictional chatter, the inverse problem formulation, and the stable and unstable regimes of the cutting process. Section.~\ref{sec3} discusses the formulation of SPINN for the forward problem and its outcomes. The forward XPINN framework is introduced, and its performance evaluation in approximating the DDE solution is presented in Section.~\ref{sec4}. Section.~\ref{sec5} introduces both the inverse SPINN and XPINN formulation and their outcomes. Finally, Section.~\ref{sec6} summarises the key outcomes of the study.


\section{Delayed Dynamics of Metal Cutting}\label{sec2}
\subsection{Metal cutting system with regenerative and frictional chatter}
\begin{figure}[!hbt]
    \centering
    \includegraphics[width=.6\linewidth]{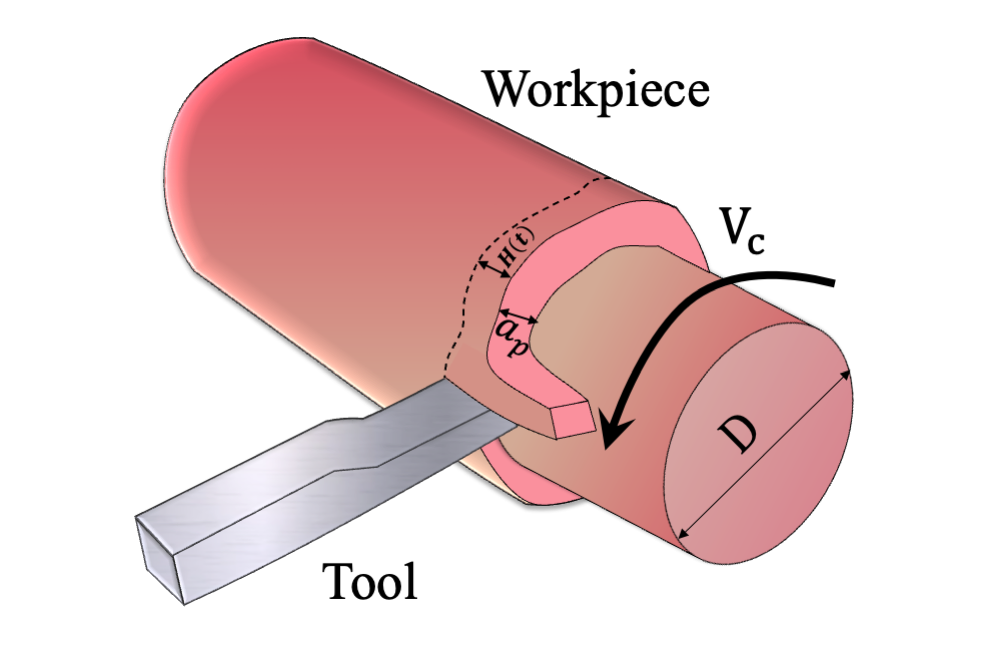}
    \caption{Schematic representation of the turning process, illustrating the relative positioning and motion of the cutting tool and workpiece.}
    \label{fig:Turning-1}
\end{figure}

\noindent In orthogonal metal cutting, a single-point cutting tool positioned perpendicular to the workpiece removes material through deformation by generating chips. Turning is an example of this orthogonal cutting mechanism in which a stationary single-point tool machines a rotating cylindrical workpiece, removing material to reduce diameter and shape external surfaces as shown in Fig.~\ref{fig:Turning-1}. The cutting process refers to the complex dynamical interaction between a machine tool and a workpiece. The interaction between the cutting process and the structural dynamics of the machine-tool system can result in self-excited vibrations known as machining chatter.

\par Chatter is an undesirable dynamic phenomenon that can significantly alter the system's stability. A major type of chatter that occurs in turning processes is regenerative chatter that arises from the interaction between the current tool position and the surface finish left by the previous revolution of the workpiece \cite{turning-wahi2008self, turning-stepan2001modelling}. Consequently, a small change in instantaneous chip thickness modifies the cutting force, altering the tool's displacement, and imprinting this deviation on the workpiece surface. Friction is another key source of chatter in cutting processes \cite{turning-wiercigroch2001frictional, turning-rusinek2014modelling}, occurring at the tool-chip and tool-workpiece contact interfaces and dependent on the relative motion and contact conditions. During turning, the friction component of the cutting force depends on static and dynamic coefficients of friction and relative velocity. Friction introduces additional nonlinear behaviours into the tool dynamics, resulting in changes in the response amplitudes and affecting the stability of oscillations.

\par The regenerative and frictional mechanisms act simultaneously. The regenerative contribution depends on the current and delayed tool states, whereas the frictional contribution depends on the instantaneous relative motion. The interaction between these mechanisms makes the cutting process fundamentally different from a simple linear vibration problem, as the resulting dynamics contain both delayed feedback and nonlinear frictional effects, and their interaction can determine whether a cutting process remains stable or develops self-excited chatter vibrations.

\subsection{Mathematical formulation of delayed cutting dynamics}\label{sec:math}

\par
\begin{figure}[!hbt]
    \centering
    \includegraphics[width=.6\linewidth]{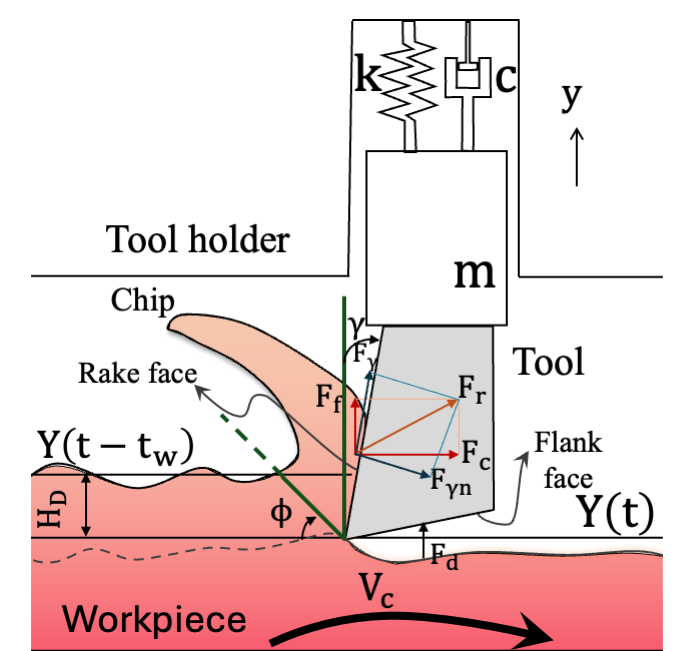}
    \caption{Illustration of the dynamic model used to describe the turning process, showing the cutting tool as an equivalent mass-spring-damper system subjected to the forces generated during the cutting process.}
    \label{fig:Turning-2}
\end{figure}

 The coupled regenerative and frictional effects described above can be formulated as a nonsmooth delayed dynamical system. The tool dynamics are represented by a classical single-degree-of-freedom (SDOF) model with mass $m\;[kg]$, damping $c\;[N\;s\;m^{-1}]$, and stiffness $k\;[N\;m^{-1}]$, as illustrated schematically in Fig.~\ref{fig:Turning-2} \cite{jaganathan2026interplay}. The workpiece has a radius $R=D/2\;[m]$ and rotates at a spindle speed of $N\;[rev\;min^{-1}]$, giving the cutting speed as $V_c=\frac{2\pi RN}{60}\;[m\;s^{-1}]$. During the turning process, chip formation occurs through elasto-plastic deformation as the stationary cutting tool engages with the rotating workpiece \cite{wiercigroch2001sources}.

During cutting, the material is sheared and flows along the rake face of the tool, which is inclined at the rake angle $\gamma$ relative to the cutting direction. The resulting chip moves along the rake face with a velocity $V_{ch}\;[m\;s^{-1}]$. As illustrated in Fig.~\ref{fig:Turning-2}, the interaction between the chip and the rake face generates the tangential force $F_{\gamma}\;[N]$ and the normal force $F_{\gamma n}\;[N]$, associated with frictional sliding and normal contact, respectively. These force components combine to produce the resultant force $F_r\;[N]$, which can be resolved into the cutting force $F_c\;[N]$ acting in the $x$-direction and the feed force $F_f\;[N]$ acting in the $y$-direction. The resulting vertical dynamics of the cutting tool, as shown in Fig.~\ref{fig:Turning-2}, are therefore governed by

\begin{equation}
    m\Ddot{Y}(t)+c\dot{Y}(t)+kY(t) = F_f + F_d,\label{eq-dim}
\end{equation}
with

\begin{equation}
    F_f = F_\gamma \cos{(\gamma)} - F_{\gamma n} \sin{(\gamma)},
\end{equation}
where $\gamma$ represents the rake angle of the cutting tool. The corresponding normal and tangential force components are expressed as follows
\begin{subequations}\label{eqn-gamma-forces}
    \begin{equation}
        F_{\gamma n} = K a_p H(t),
    \end{equation}
    \begin{equation}
        F_{\gamma} = \mu F_{\gamma n},
    \end{equation}
\end{subequations}
where $a_p$ denotes the depth of cut, and $H(t)$ represents the instantaneous chip thickness, which is given by the following expression
\begin{equation}
    H(t) = H_D - Y(t) + Y(t-t_w).
\end{equation}
Here, $H_D\;[m]$ denotes the nominal chip thickness (feed rate), while $t_w=\frac{60}{N}\;[s]$ represents the rotational period of the workpiece. The parameter $K$ in Eq.~\eqref{eqn-gamma-forces} denotes the cutting force coefficient, and $\mu$ represents the equivalent friction coefficient associated with the Stribeck friction model \cite{turning-bai2017improved}.

\begin{equation}
    \mu = \mathrm{sign}(V_\gamma)\left(\mu_d + (\mu_d-\mu_s)\exp{\left(-\dfrac{|V_\gamma|}{V_s}\right)}\right),
\end{equation}
where $V_s\;[m\;s^{-1}]$ denotes the Stribeck velocity, while $\mu_d$ and $\mu_s$ represent the dynamic and static friction coefficients, respectively. The quantity $V_\gamma\;[m\;s^{-1}]$ denotes the relative frictional velocity between the chip and the tool and is given by

\begin{equation}
    V_\gamma = V_{ch}-\dot{Y}(t)\cos(\gamma),
\end{equation}
where $V_{ch} = V_c\frac{\sin{(\phi)}}{\cos{(\phi-\gamma)}}\;[m\;s^{-1}]$ denotes the chip velocity and $\phi$ denotes the shear angle of the cutting tool. 

\par $F_d$ in Eq.~\eqref{eq-dim} denotes the process damping force, and it provides an important stabilising mechanism in machining dynamics. Process damping arises from the interaction between the vibrating tool flank face and the uncut workpiece material. Its contribution can be represented through a velocity-dependent damping force, introducing an effective process damping coefficient $C_y$ into the governing dynamics, and is given as
\begin{equation}\label{eq-process_damping}
        F_d = - C_y a_p \dfrac{\dot{Y}(t)}{V_c},
\end{equation}

Introducing suitable non-dimensional variables into Eq.~\ref{eq-dim} yields the following non-dimensional turning model:

\begin{subequations}\label{non-dim-eqs}
    \begin{equation}
        \xi=\dfrac{c}{\sqrt{mk}}, \quad v_s=\dfrac{30V_s \cos{(\gamma-\phi)}}{\pi R \sin{(\phi)}}\sqrt{\dfrac{m}{k}},
    \end{equation}
    \begin{equation}
        c_y = \dfrac{30C_y}{\pi R K}, \quad \nu=\dfrac{H_D}{V_s}\sqrt{\dfrac{k}{m}}, \quad n=N\sqrt{\dfrac{m}{k}},
    \end{equation}
    \begin{equation}\label{eq-nondim-depth}
        W = a_p \dfrac{K}{k}, 
    \end{equation}
    \begin{equation}
        \tau_w=\dfrac{60}{n}=t_w\sqrt{\dfrac{k}{m}}, \quad \Omega=\dfrac{2\pi}{\tau_w}
    \end{equation}
    \begin{equation}
        \tau=t\sqrt{\dfrac{k}{m}}, 
    \end{equation}
    \begin{equation}\label{Eq-nondimY}
        y_1(\tau) = \dfrac{Y(t)}{H_D},
    \end{equation}
    \begin{equation}
        y_2(\tau)=\dot{y}_1(\tau)=\dfrac{\dot{Y}(t)}{H_D}\sqrt{\dfrac{m}{k}},
    \end{equation}
    \begin{equation}
        v_\gamma = \dfrac{V_\gamma}{H_D}\sqrt{\dfrac{m}{k}} = \dfrac{n}{v_s}-\nu \cos(\gamma) y_2(\tau).
    \end{equation}
\end{subequations}
The corresponding non-dimensional form of the model is given by

\begin{subequations}
    \begin{equation}
        \ddot{y}(\tau)+\xi \dot{y}(\tau) + y(\tau) = f_f +f_d, \label{eq-nondim}
    \end{equation}
with 
    \begin{equation}
        f_f = W (\mu_{vs}\cos{(\gamma)-\sin{(\gamma)}}) h(\tau), \label{eq:ff}
    \end{equation}
    \begin{equation}
        f_d = - W c_y \dfrac{\dot{y}}{n}, \label{eq:fd}
    \end{equation}
    \begin{equation}\label{eq:frictional}
        \mu_{vs} = \mathrm{sign}(g)\left(\mu_d+(\mu_s-\mu_d)e^{|-g|}\right),
    \end{equation}
    \begin{equation}
  \text{where} \,\,\,  g = \frac{n}{v_s(\gamma,\phi)}-\nu \cos (\gamma) \dot{y}(\tau),
    \end{equation}
    \begin{equation}
     \text{and} \, \,\,\,  h(\tau)=1-y(\tau)+y(\tau-\tau_w).
    \end{equation}
\end{subequations}

\subsection{Unknown parameters and inverse-problem formulation}
\par The mathematical formulation of the turning process presented in the previous section (Sec.~\ref{sec:math}) contains various physical and operating parameters that characterise the machining dynamics. In this study, the cutting force coefficient $K$ and the process damping coefficient $C_y$ are selected as the unknown parameters to be estimated using the inverse PINN methods. The remaining parameters are assumed to be known from the machining configuration and experimental conditions.

\par The choice of $K$ and $C_y$ is motivated by their influence on the dynamics of the machining system and also by the practical challenges associated with their experimental identification. The cutting force coefficient $K$ determines the magnitude of the cutting-force contribution through the dimensionless parameter 
presented in Eq.~\eqref{eq-nondim-depth} and consequently has a direct influence on the cutting force and the resulting tool vibrations. Cutting force coefficients are typically identified from cutting-force measurements across a range of cutting conditions \cite{altintas2008identification, qiu2018modeling}. However, the cutting force coefficient depends on the tool-workpiece combination, cutting conditions, and tool geometry, so it generally requires process-specific experimental calibration rather than being a universal material constant. 

\par The process damping coefficient $C_y$ is chosen since the process damping force represents an important source of energy dissipation arising from the flank face of the tool and the uncut workpiece material. Its magnitude is affected by factors such as cutting speed, flank-face interaction, and tool wear. Budak and Tunc reported that process damping had historically been difficult to quantify due to the complexity of directly measuring the associated damping forces and proposed a procedure based on experimentally observed stability limits \cite{budak2010identification}, while Kurata et al. subsequently investigated in-process identification of process damping during turning and milling \cite{kurata2010chatter}. Altintas et al. obtained the values of process-damping coefficients using controlled orthogonal cutting experiments with a fast tool servo and piezoelectric excitation \cite{altintas2008identification}. Mechanisms and modelling involved in machining process damping are summarised in the recent review article by Feng and Lie \cite{feng2023mechanism}. The process damping force given in Eq.~\eqref{eq-process_damping} explicitly depends on the velocity of the tool. Thus, the estimation of $C_y$ requires identification of a parameter embedded within the nonlinear dynamics response rather than a simple direct measurement of a damping force. The selection of $K$ and $C_y$ is also motivated by their distinct roles in the present non-dimensional model. Specifically,
\begin{equation}
     W = a_p \dfrac{K}{k}, \qquad c_y = \dfrac{30C_y}{\pi R K},
\end{equation}
The cutting force coefficient $K$ controls the scaling of the cutting force through $W$, and $C_y$ contributes to the velocity-dependent process damping through $c_y$. Also, $K$ appears in the denominator of $c_y$, resulting in a coupling between the two unknown parameters. Therefore, the estimation of $K$ and $C_y$ cannot be considered as two independent parameter-fitting problems. Instead, they become a coupled nonlinear parameter-estimation problem in which modifications in any single parameter simultaneously affect the multiple terms in the governing DDE. This makes $K$ and $C_y$ particularly suitable candidates for physics-informed inverse estimation, where the governing machining dynamics and measured response can be used simultaneously to infer their values.

\subsection{Stable and Unstable cases}\label{sec2pt4}
\par The dynamic response of the considered machining system depends on the selected cutting conditions, particularly the spindle speed ($N$) and the depth of the cut ($a_p$). Depending on these operating conditions, the tool vibrations in cutting dynamics can either die down or lead to self-excited oscillations. A stable cutting condition corresponds to a chatter-free response of the system, whereas an unstable cutting condition corresponds to the development of chatter, characterised by growing oscillations in the dynamic response.
\begin{figure}[!hbt]
    \centering
    \includegraphics[width=.8\linewidth]{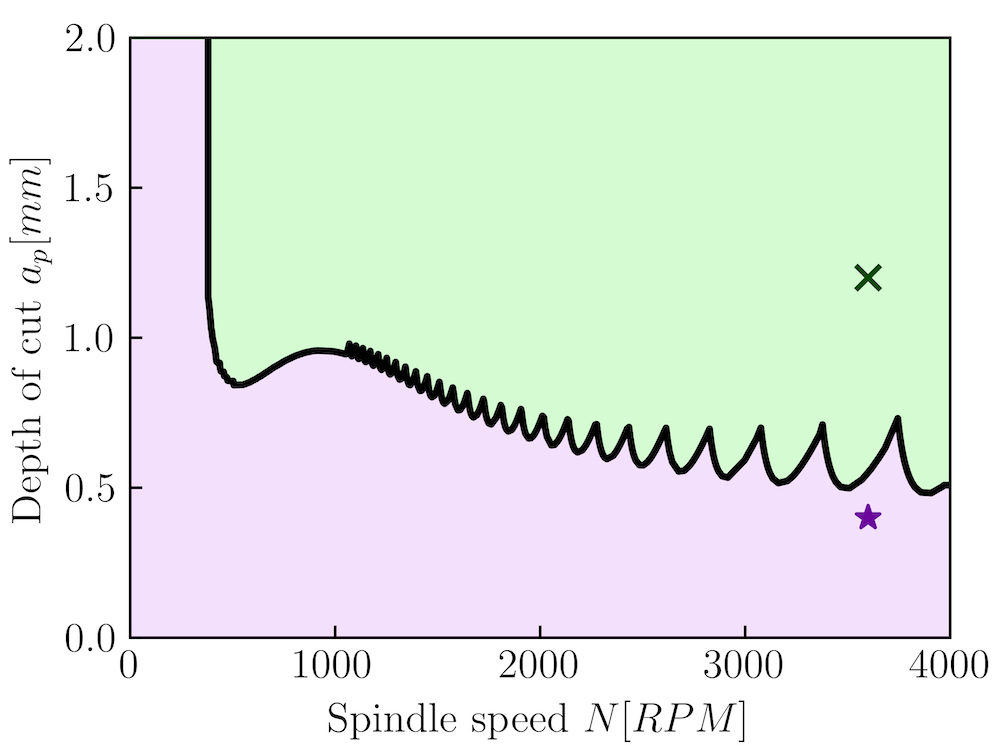}
    \caption{Stability lobe diagram showing the stable (shaded in purple) and unstable (shaded in green) operating regimes, with the selected parameters for the stable and unstable cutting cases indicated by a star and a cross, respectively. The stable cutting condition corresponds to a depth of cut of $a_p=0.4$ mm, while the unstable (chatter) cutting condition corresponds to $a_p=1.2$ mm, both at a spindle speed of $N=3600$ RPM.}
    \label{fig:SLD}
\end{figure}
\par To illustrate the two operating regimes considered in this study, the stability characteristics of the governing DDE are represented using a stability lobe diagram (SLD) in the $a_p$-$N$ parameter space, as shown in Fig.~\ref{fig:SLD}. The computational steps involved in getting this SLD for the considered DDE can be found in Refs.~\cite{jaganathan2026interplay, turning-yan2019modelling}. The black line in Fig.~\ref{fig:SLD} represents the stability boundary, and it separates the stable (shaded in purple) and unstable (shaded in green) operating regimes. 

\par For the subsequent analysis, two representative operating parameters are selected from the SLD, one in the stable cutting regime and another in the unstable one, denoted by a star and cross mark, respectively; See Fig.~\ref{fig:SLD}. Considering both cases is important for evaluating the ability of the proposed PINN frameworks to represent different dynamic regimes of the machining system. Furthermore, using both stable and unstable operating conditions provides a more comprehensive assessment of the inverse PINN framework for estimating the unknown machining parameters $K$ and $C_y$. The parameter-estimation performance is therefore evaluated using response data generated under both regimes, allowing the proposed SPINN and XPINN formulations to be assessed not only for chatter-free stable cutting but also for the more challenging chatter condition.

\section{Forward SPINN Formulation for solving machining dynamics}\label{sec3}
\par This section introduces the forward single-domain PINN method for solving the DDE given by Eq.~\eqref{eq-dim} and discusses its performance. The forward PINN frameworks are introduced first because parameter estimation with PINNs requires a reliable, accurate representation of the system's underlying dynamics.
\subsection{Single-domain PINN formulation}
\par
\begin{figure}[!hbt]
    \centering
    \includegraphics[width=.8\linewidth]{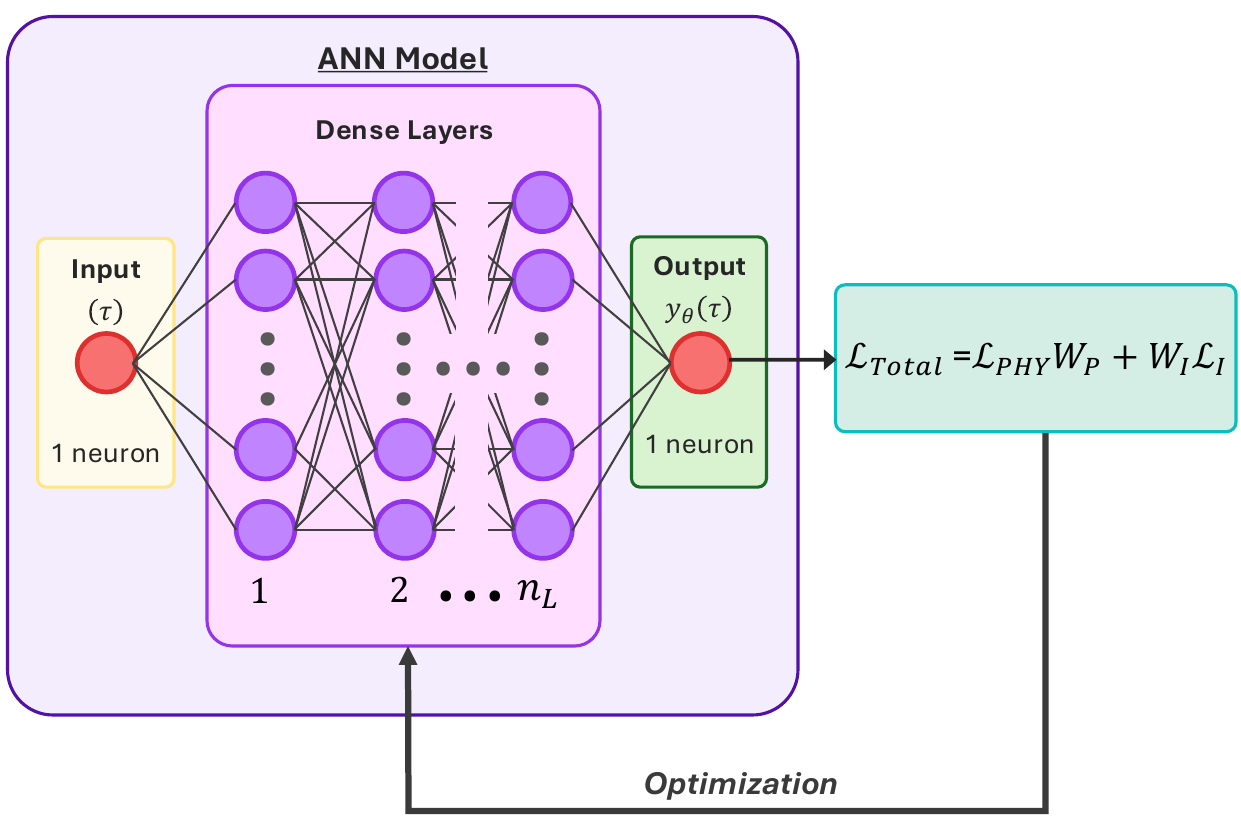}
    \caption{Schematic of the PINN for solving the forward problem}
    \label{fig:sche-spinn-for}
\end{figure}

 The schematic diagram of the single-domain PINN (SPINN) framework is presented in Fig.~\ref{fig:sche-spinn-for}. The Artificial Neural Network (ANN) with one input layer, one output layer, and $n_L$ hidden layers approximates the solution of the DDE, say $y_\theta(\tau)$, for the given non-dimensional time $\tau$. Let the ANN $(\mathcal{N}_\theta)$ approximation be denoted by
\begin{equation}
    y_\theta(\tau) = \mathcal{N}_\theta(\tau),
\end{equation}
where $\theta$ represents the trainable parameters of $\mathcal{N}_\theta$. For a given input $\tau$, the solution $y(\tau)$, its first derivative $\dot{y}(\tau)$, and its second derivative $\ddot{y}(\tau)$ are computed as 
\begin{equation}
y_\theta(\tau) = y_\theta(\tau), \qquad
\dot{y}_\theta(\tau) = \frac{\mathrm{d}y_\theta}{\mathrm{d}\tau}, \qquad
\ddot{y}_\theta(\tau) = \frac{\mathrm{d}^2y_\theta}{\mathrm{d}\tau^2},
\end{equation}
efficiently using Automatic Differentiation (AD) \cite{baydin2018automatic, margossian2019review} by applying the chain rule. Backpropagation \cite{backproprumelhart1986learning} in ANNs is implemented using the AD principle to propagate gradients through the network and compute the derivatives of the output with respect to the network's parameters and inputs. The approximation of the state variable was carried out using a fully connected feedforward neural network with a hyperbolic tangent activation function as 
\begin{equation}
    Z_l = \tanh{w_l.Z_{l-1}+b_l}, \qquad l=1,2,3,\dots,n_L,
\end{equation}
where $Z_l\in\mathbb{R}^{d_l}$ denotes the output of the $l^{\mathrm{th}}$ layer, $w_l\in\mathbb{R}^{d_l\times d_{l-1}}$ is the corresponding weight matrix, and
$b_l\in\mathbb{R}^{d_l}$ is the bias vector. Here,
$d_l=\dim(Z_l)$, and the hyperbolic tangent function is applied element-wise. Throughout this work, PyTorch \cite{paszke2019pytorch} is used to implement backpropagation, AD, and other optimisation steps involved in the PINN frameworks. 
The network parameters, consisting of the weights and biases, are optimised by minimising a loss function. For the case of SPINN, the total loss function $\mathcal{L}_{{Total}}$ is composed of the physical loss $\mathcal{L}_{PHY}$ constructed from the governing DDE and the initial condition loss function $\mathcal{L}_I$. The physics loss is defined as the mean square residual over the collocation set $\{\tau_i\}_{i=1}^{N_f}$:

\begin{equation}\label{Eq:phy_loss}
\mathcal{L}_{PHY} = \frac{1}{N_f}\sum_{i=1}^{N_f}\left|\mathcal{R}(\tau_i)\right|^2,
\end{equation}
where $N_f$ is the number of collocation points and $\mathcal{R}(\tau_i)$ is the residual value of the considered DDE. By rewriting Eq.~\eqref{eq-dim}, the residual at any $\tau$ is expressed as
\begin{equation}\label{Eq:residualDDE}
    \mathcal{R}(\tau;\theta) = \ddot{y}_\theta(\tau)+\xi \dot{y}_\theta(\tau) + y_\theta(\tau) - f_f - f_d.
\end{equation}
The delayed state is evaluated using a piecewise rule, i.e., if  $\tau > \tau_w$, the delayed value is computed from the network as
\begin{equation}
y(\tau-\tau_w) = y_\theta(\tau-\tau_w),
\end{equation}
while for $\tau \leq \tau_w$ the delayed term is replaced by a history function 
\begin{equation}
y(\tau-\tau_w) = \Phi_{H}(\tau), \qquad \tau \leq \tau_w.
\end{equation}
The initial condition loss is computed as \cite{wang2025deep}
\begin{equation}
    \mathcal{L}_I = \left|y_\theta(0)-\Phi_H(0)\right|^2.
\end{equation}
Therefore, the total loss is 
\begin{equation}
    \mathcal{L}_{Total} = W_{P} \mathcal{L}_{PHY} + W_I \mathcal{L}_I,
\end{equation}
where 
\begin{equation}
    W_{P} = \dfrac{\mathcal{L}_{PHY}}{\mathcal{L}_{PHY}+\mathcal{L}_I}, \qquad W_I =  \dfrac{\mathcal{L}_{I}}{\mathcal{L}_{PHY}+\mathcal{L}_I},
\end{equation}
are the adaptive weights used to normalise the individual losses during training.

\subsection {Data-driven solution of machining system using SPINNs}
\par The training data for solving the considered DDE consists of $N_f = 5\times10^4$ collocation points sampled uniformly over the interval $[\tau_{min}, \tau_{max}]$. The number of hidden layers $n_L$ is set to $5$, and the ANN's architecture is taken to be $[1, 32, 128, 128, 128, 32, 1]$, where the input dimension is one temporal variable, and the output is the scalar approximation of the system response. As mentioned earlier, all hidden layers use the $\tanh$ activation function. The ANN is optimised using the Adam \cite{kingma2014adam} optimiser, which updates the weights and biases to minimise the total loss $\mathcal{L}_{Total}$. The initial learning rate is set to $10^{-3}$. An exponential learning-rate scheduler was applied during training with a decay factor
$\gamma_{\mathrm{lr}} = 0.9^{1/10000}$, resulting in a gradual reduction of the learning rate at each training epoch and a cumulative $10\%$ reduction over $10^{4}$ epochs. The model is trained for a total of $2\times10^5$ epochs. At each iteration, the residual-based loss is computed, backpropagation is performed, and the optimiser updates the network parameters. This process is repeated until the final training epoch. After training, the network predictions are evaluated over the full temporal grid. Let $y_{\mathrm{ref}}(\tau)$ denote the reference solution extracted from MATLAB using the DDE23 solver \cite{shampine2001solving} and $y_{pre}(\tau)$ the PINN prediction. The relative $L_2$ error was computed as
\begin{equation}
L_2 = \frac{\|y_{\mathrm{pre}} - y_{\mathrm{ref}}\|_2}{\|y_{\mathrm{ref}}\|_2}.
\end{equation}
The Mean Absolute Error (MAE) was also calculated as
\begin{equation}
MAE = \dfrac{1}{N_f}\sum_{i=1}^{N_f}\left|y_{\mathrm{pre}} - y_{\mathrm{ref}}\right|
\end{equation}
\begin{figure}[!hbt]
    \centering
    \includegraphics[width=\linewidth]{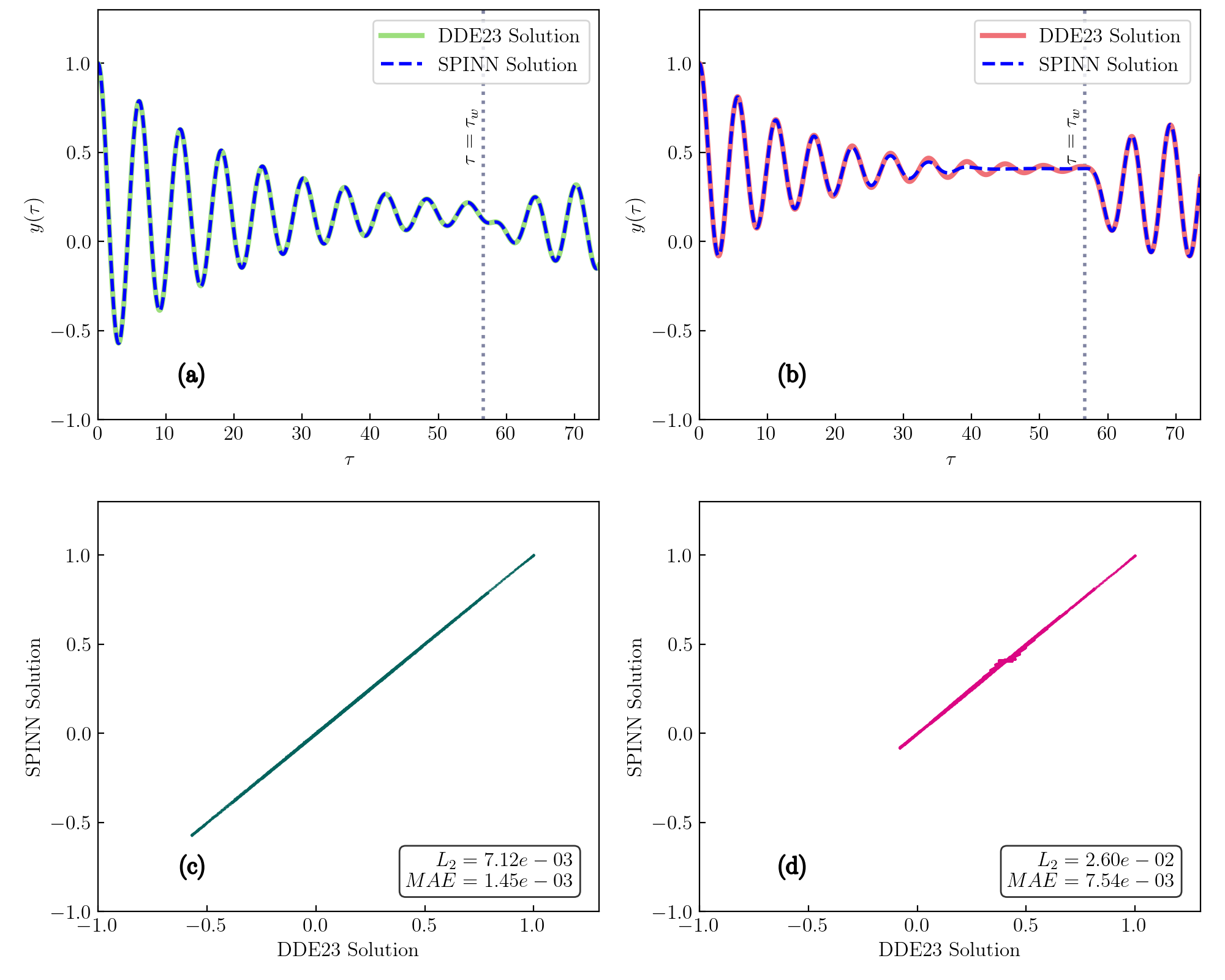}
    \caption{SPINN performance in solving the machining system with a constant history function $\Phi_H=1$. (a) and (b) represent plots of predicted values (dotted line) over the DDE23 reference solution (solid line) for the stable and unstable cases, respectively. (c) and (d) correspond to the scatter plots of the DDE solution with the PINN solution, respectively, for the stable and unstable cases.}
    \label{fig:spinn-const-small}
\end{figure}
For the results presented in this paper, the analyses are performed for two primary cases, distinguished by the depth of cut $a_p$: (i) a stable cutting condition ($a_p=0.4\; mm$) and (ii) an unstable or chatter cutting condition ($a_p=1.2\;mm$). All remaining parameters are fixed and adopted from \cite{jaganathan2026interplay, turning-yan2019modelling}, with $N=3600\; RPM$, $m=0.561\;[kg]$, $c = 145\;[Nsm^{-1}]$, $k = 6.48\times 10^6\;[Nm^{-1}]$, $R = 0.0175 [m]$, $H_D = 0.0005\;[m]$, $\gamma = 0^o$, $\phi = 45^o$, $V_s = 0.65\;[ms^{-1}]$, $\mu_d = 0.23$, $\mu_s = 0.54$. Also, the values of the cutting force coefficient and the process damping coefficient are taken as $K = 6.02\times 10^9\;[Nm^{-2}]$, $C_y = 6.11\times 10^5\;[Nm^{-1}]$. These coefficients ($K, C_y$) are treated as the ground-truth set for the inverse problem discussed in the subsequent sections.
\begin{figure}[!hbt]
    \centering
    \includegraphics[width=\linewidth]{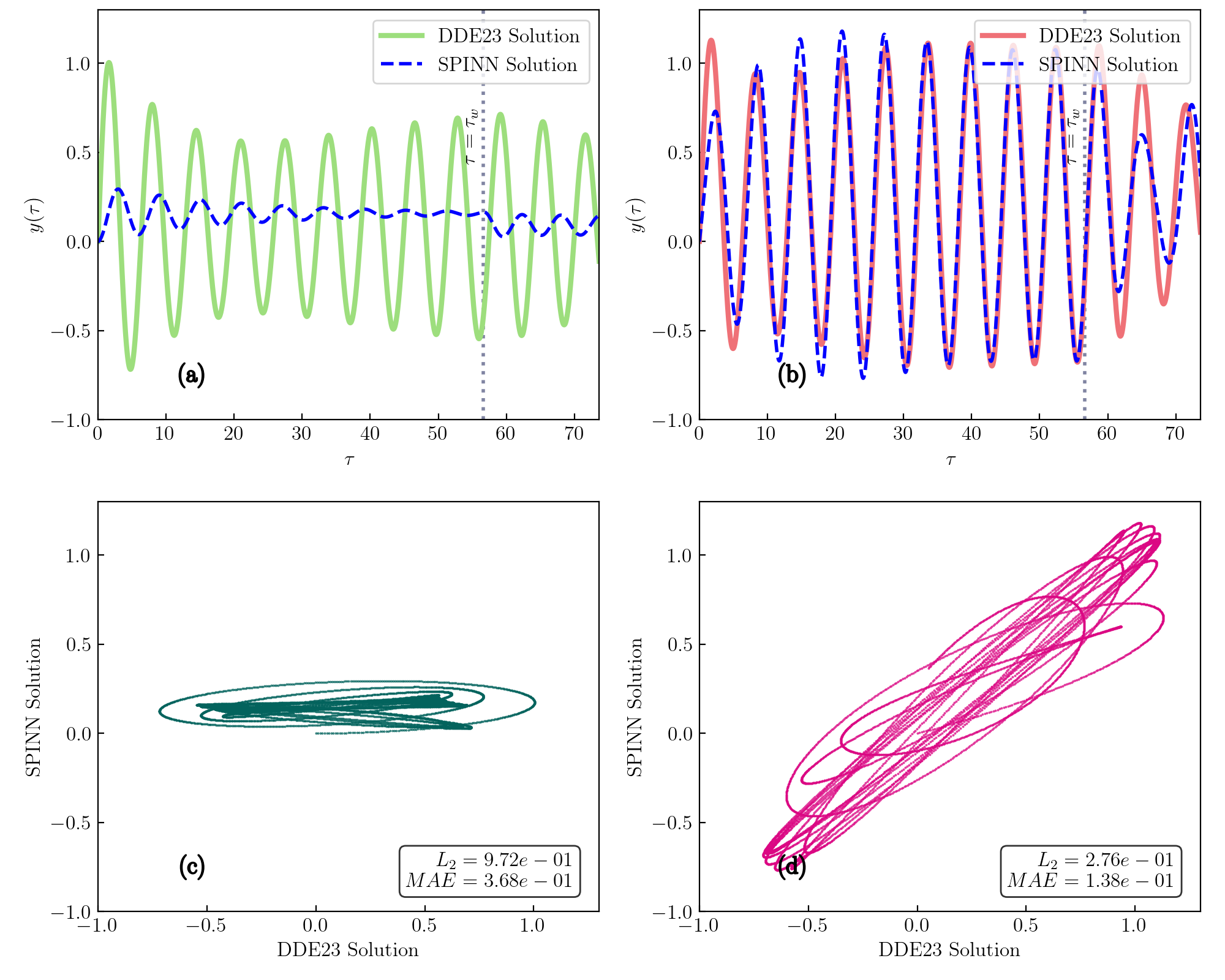}
    \caption{SPINN performance in solving the machining system with a history function $\Phi_H(\tau)=\sin(\tau)$. (a) and (b) represent plots of predicted values (dotted line) over the DDE23 reference solution (solid line) for the stable and unstable cases, respectively. (c) and (d) correspond to the scatter plots of the DDE solution with the PINN solution, respectively, for the stable and unstable cases.}
    \label{fig:spinn-sin-small}
\end{figure}
\par First, the analysis is carried out for a simulation time of $\tau\in[0, 1.3\tau_w]$, slightly more than the delay value $\tau_w$, with a constant history function of $\Phi_H=1$. After training, the SPINN framework's performance is evaluated by predicting the solution for the given time. The SPINN results for simulation time up to $1.3\tau_w$ are represented in Fig.~\ref{fig:spinn-const-small}. Fig.~\ref{fig:spinn-const-small}~(a)\&(b) show that the SPINN is able to approximate the solution of the considered DDE \eqref{eq-dim}. The straight-line behaviour in scattered plots presented in Fig.~\ref{fig:spinn-const-small}~(c)\&(d) also confirms the accurate approximation of SPINN for both stable and unstable cases. The relative $L_2$ error values are $7.12\times10^{-3}$ and $2.6\times10^{-2}$ and the MAEs are calculated as $1.45\times10^{-3}$ and $7.54\times10^{-3}$ for stable and unstable cases respectively.
\begin{figure}[!hbt]
    \centering
    \includegraphics[width=\linewidth]{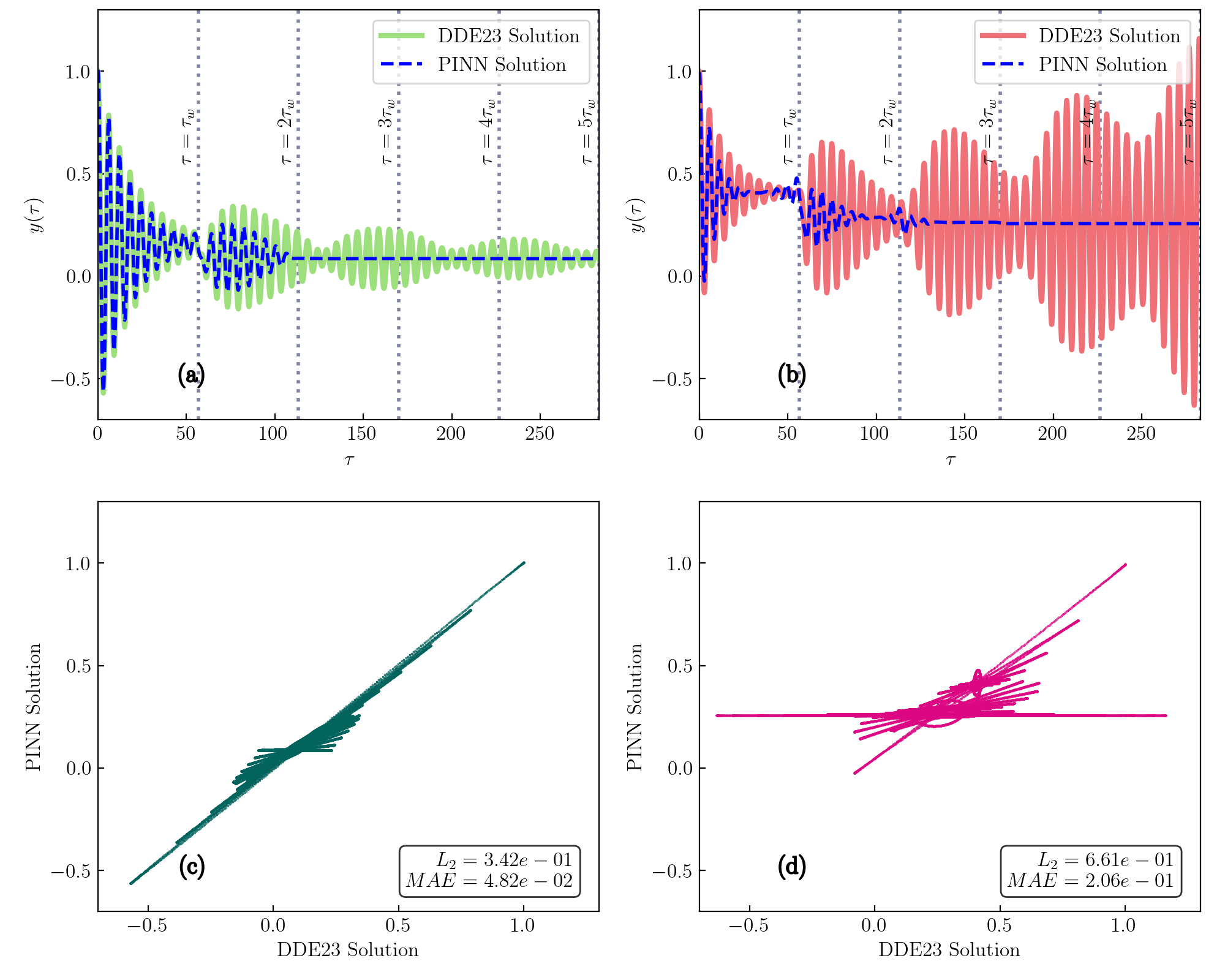}
    \caption{SPINN results in solving the machining system for simulation time $\tau\in[0, 5\tau_w]$ with a constant history function $\Phi_H(\tau)=1$ and $N_f=5\times10^4$ sampling points. (a) and (b) represent plots of predicted values (dotted line) over the DDE23 reference solution (solid line) for the stable and unstable cases, respectively. (c) and (d) correspond to the scatter plots of DDE solution with PINN solution, respectively, for stable and unstable cases.}
    \label{fig:spinn-const-long}
\end{figure}
\par Next, the same analysis is repeated with a non-constant history function, specifically for $\Phi_H(\tau)=\sin(\tau)$, and the corresponding results are presented in Fig.~\ref{fig:spinn-sin-small}. Figs.~\ref{fig:spinn-sin-small}~(a)\&(c) infer that the performance of approximating the solution for the stable case is poor, with the relative $L_2$ error value of $9.72\times10^{-1}$ and $MAE$ value of $3.68\times10^{-1}$. For the unstable case, the scatter plots in Figs.~\ref{fig:spinn-sin-small}~(b)\&(d) show that the SPINN fails to approximate the solution of the considered DDE \eqref{eq-dim} with the error values of $L_2=2.76\times10^{-1}$ and $MAE=1.38\times10^{-1}$.
\begin{figure}[!hbt]
    \centering
    \includegraphics[width=\linewidth]{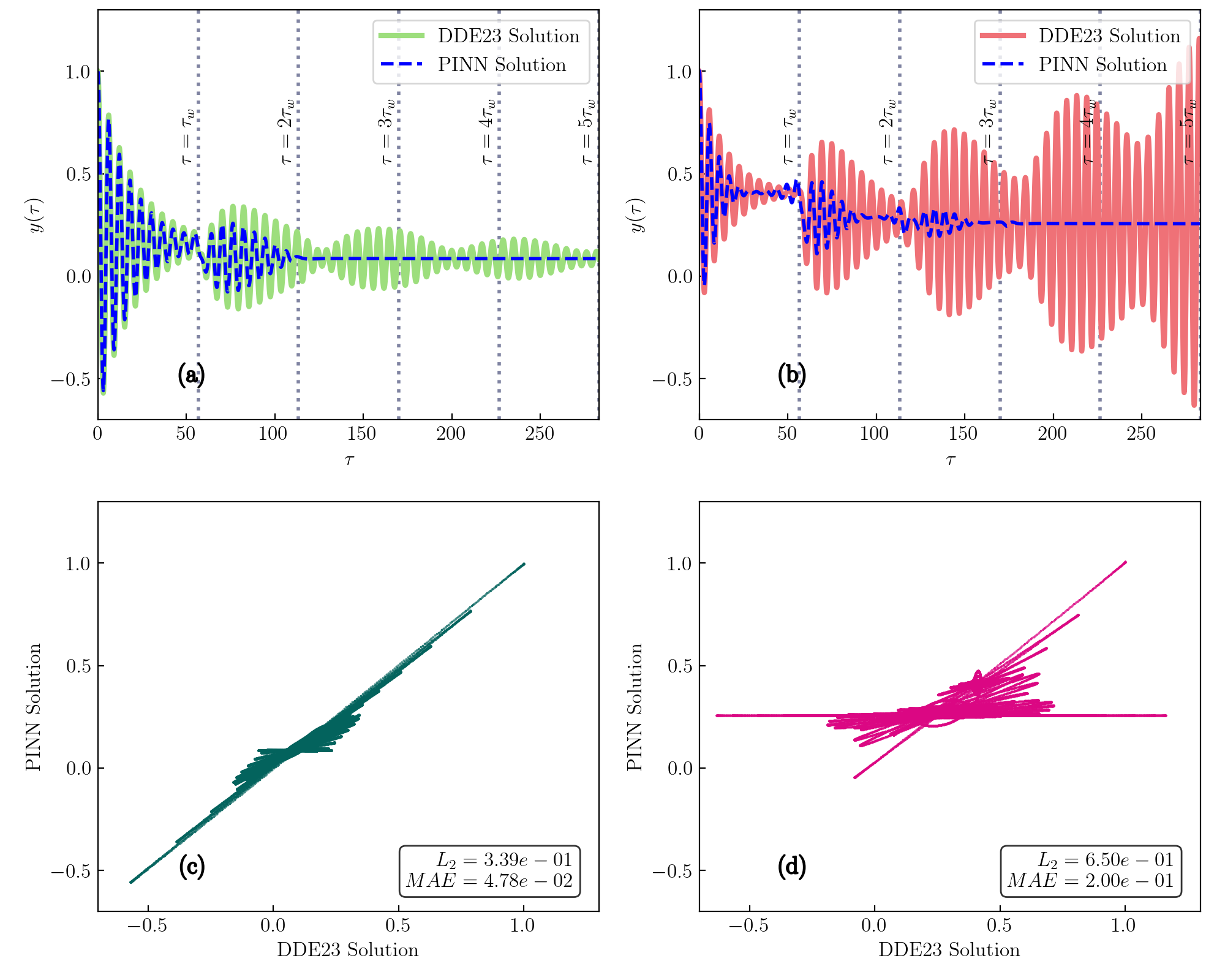}
    \caption{SPINN results in solving the machining system for simulation time $\tau\in[0, 5\tau_w]$ with a constant history function $\Phi_H(\tau)=1$ and $N_f=25\times10^4$ sampling points. (a) and (b) represent plots of predicted values (dotted line) over the DDE23 reference solution (solid line) for the stable and unstable cases, respectively. (c) and (d) correspond to the scatter plots of the DDE solution with the PINN solution, respectively, for the stable and unstable cases.}
    \label{fig:spinn-const-long2}
\end{figure}
\par To test the ability of SPINN in approximating the DDE solution for longer simulation time, the proposed SPINN framework experimented with a constant history function $\Phi_H(\tau)=1$ for $\tau\in[0, 5\tau_w]$, and the results are presented in Fig.~\ref{fig:spinn-const-long}. In this case, the number of sampling points $N_f$ is fixed at $5\times10^4$, as in the previous two cases. From Fig.~\ref{fig:spinn-const-long}~(a)\&(b), it is clear that the SPINN struggles to maintain an accurate approximation of the solution of the given machining DDE once the simulation time extends beyond twice the lag value $(\tau>2\tau_w)$. The scattered points deviating from the straight line in Fig.~\ref{fig:spinn-const-long}~(c)\&(d) further confirm that the SPINN fails to capture the underlying dynamics of the machining DDE in the considered regime.

\par
\begin{figure}[!hbt]
    \centering
    \includegraphics[width=\linewidth]{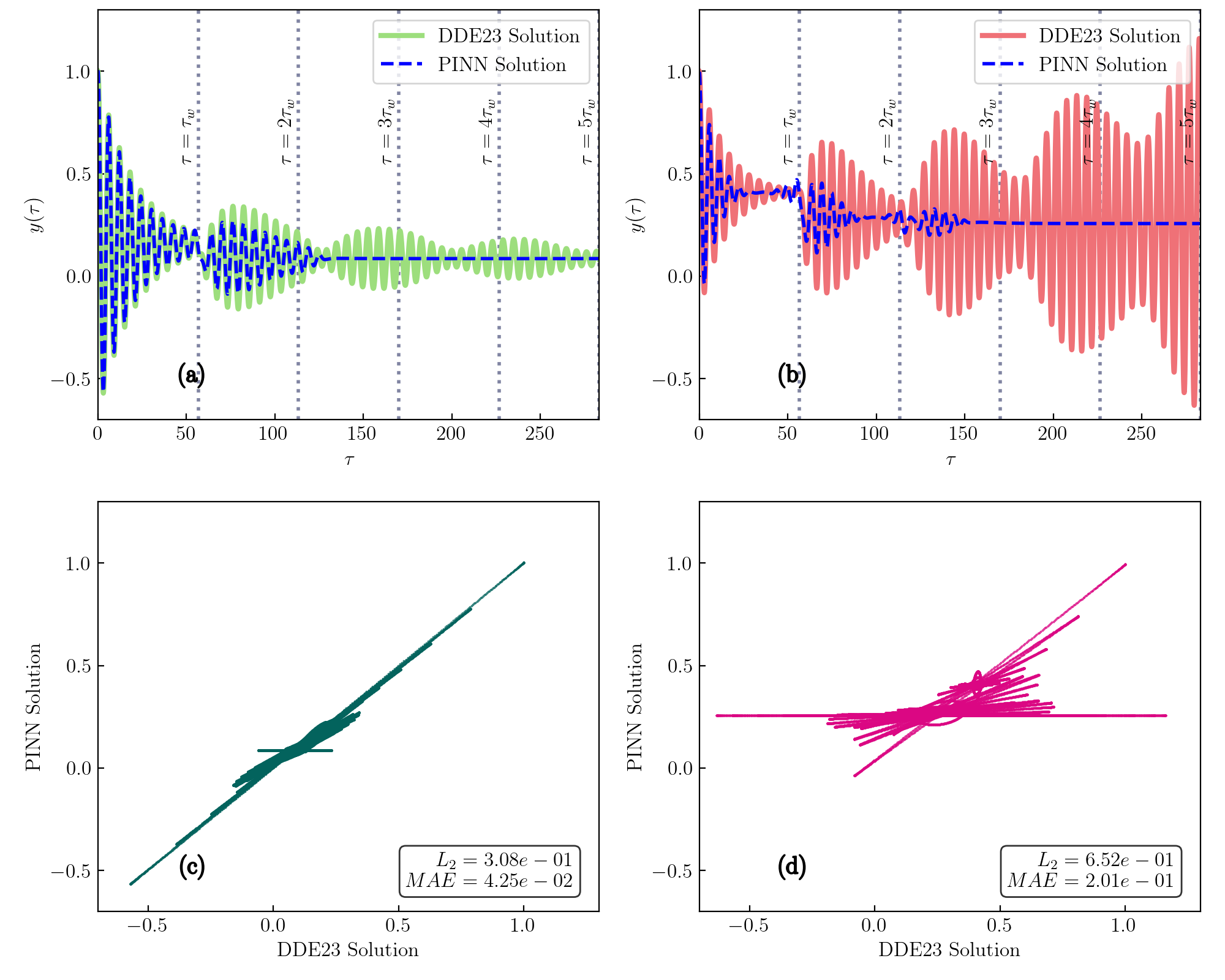}
    \caption{SPINN results in solving the smooth approximated machining system for simulation time $\tau\in[0, 5\tau_w]$ with a constant history function $\Phi_H(\tau)=1$ and $N_f=25\times10^4$ sampling points. (a) and (b) represent plots of predicted values (dotted line) over the DDE23 reference solution (solid line) for the stable and unstable cases, respectively. (c) and (d) correspond to the scatter plots of the DDE solution with the PINN solution, respectively, for the stable and unstable cases.}
    \label{fig:spinn-const-long2_smooth}
\end{figure}

 In the first analysis over  $\tau\in[0, 1.3\tau_w]$, a considerably large number of collocation points $N_f=5\times10^4$ was set for this study. The same value was used for the extended interval of $\tau \in [0, 5\tau_w]$, but the resulting performance was noticeably inadequate. To ensure a fair comparison, $N_f$ was increased to $25\times 10^4$ for the longer-time simulation, and the experiment was repeated. The updated results are shown in Fig.~\ref{fig:spinn-const-long2}. However, even with this substantial increase in sampling density, SPINN still fails to give a reliable approximation. Next, the nonsmooth frictional term in Eq.~\eqref{eq:frictional} is replaced by a smooth approximation since it introduces additional difficulties over the extended time horizons. The sign function in Eq.~\eqref{eq:frictional} is replaced with a smooth approximation through the hyperbolic tangent function,
\begin{equation}\label{eq:smooth}
\operatorname{sign}(x) \approx \tanh(1000 x).
\end{equation}
The SPINN results with this smooth approximation are shown in Fig.~\ref{fig:spinn-const-long2_smooth}. It is clear from the plots shown in Fig.~\ref{fig:spinn-const-long2_smooth} that the SPINN fails to approximate the solution even with the smooth approximation. 
\begin{table*}[h!]
\centering
\begin{tabular}{c c c c c}
\toprule
\multirow{2}{*}{$N_f$} & \multicolumn{2}{c}{$L_2$} & \multicolumn{2}{c}{$\text{MAE}$} \\
\cmidrule(lr){2-3} \cmidrule(lr){4-5}
 & Stable & Unstable & Stable & Unstable \\
\midrule
$5\times10^4$  & $3.42\times10^{-1}$ & $6.61\times10^{-1}$ & $4.82\times10^{-2}$ & $2.06\times10^{-1}$ \\
$25\times10^4 $& $3.39\times10^{-1}$ & $6.5\times10^{-1}$ &  $4.78\times10^{-2}$ & $2\times10^{-1}$ \\
$25\times10^4$ (smooth) & $3.08\times10^{-1}$ & $6.52\times10^{-1}$ &  $4.25\times10^{-2}$ & $2.01\times10^{-1}$ \\
\bottomrule
\end{tabular}
\caption{Comparison of SPINN error metrics for different numbers of collocation points over the interval $\tau\in[0,5\tau_w]$ and a smooth approximation of the nonsmooth friction with a constant history function.}
\label{tab:error_compar_spinn_5tau}
\end{table*}
Also, Table~\ref{tab:error_compar_spinn_5tau} shows that increasing the number of collocation points from $N_f=5\times10^4$ to $25\times10^4$ and also a smooth approximation produce only marginal improvements in both $L_2$ and $MAE$ errors. This indicates that the SPINN exhibits significant difficulty in approximating the solution of the machining DDE once the simulation time exceeds twice the lag value.

The above SPINN experiments for solving the DDE governing the machining dynamics reveal the two major limitations:
\begin{itemize}
    \item[(i)] SPINN fails to approximate the solution to the DDE when a non-constant history function is considered.
    \item[(ii)] SPINN struggles to maintain accuracy in solving the considered DDE with an extended time interval beyond $\tau_w$.
\end{itemize}
In machining models, particularly in turning operations, the initial waviness profile of the workpiece is commonly represented via superposition of harmonic functions \cite{jaganathan2026interplay, yan2017basins}. Moreover, accurate long-time simulation is essential for determining whether the chatter vibrations will emerge and persist throughout the cutting process. It necessitates a framework that reliably approximates the DDE solution for both non-constant history functions and extended simulation time horizons. The next section introduces the XPINN framework, identifies the shortcomings of SPINN, and presents the modifications needed to address these challenges.


\section{Forward XPINN formulation for solving machining dynamics}\label{sec4}
\begin{figure*}[!hbt]
        \centering
        \includegraphics[width=\linewidth]{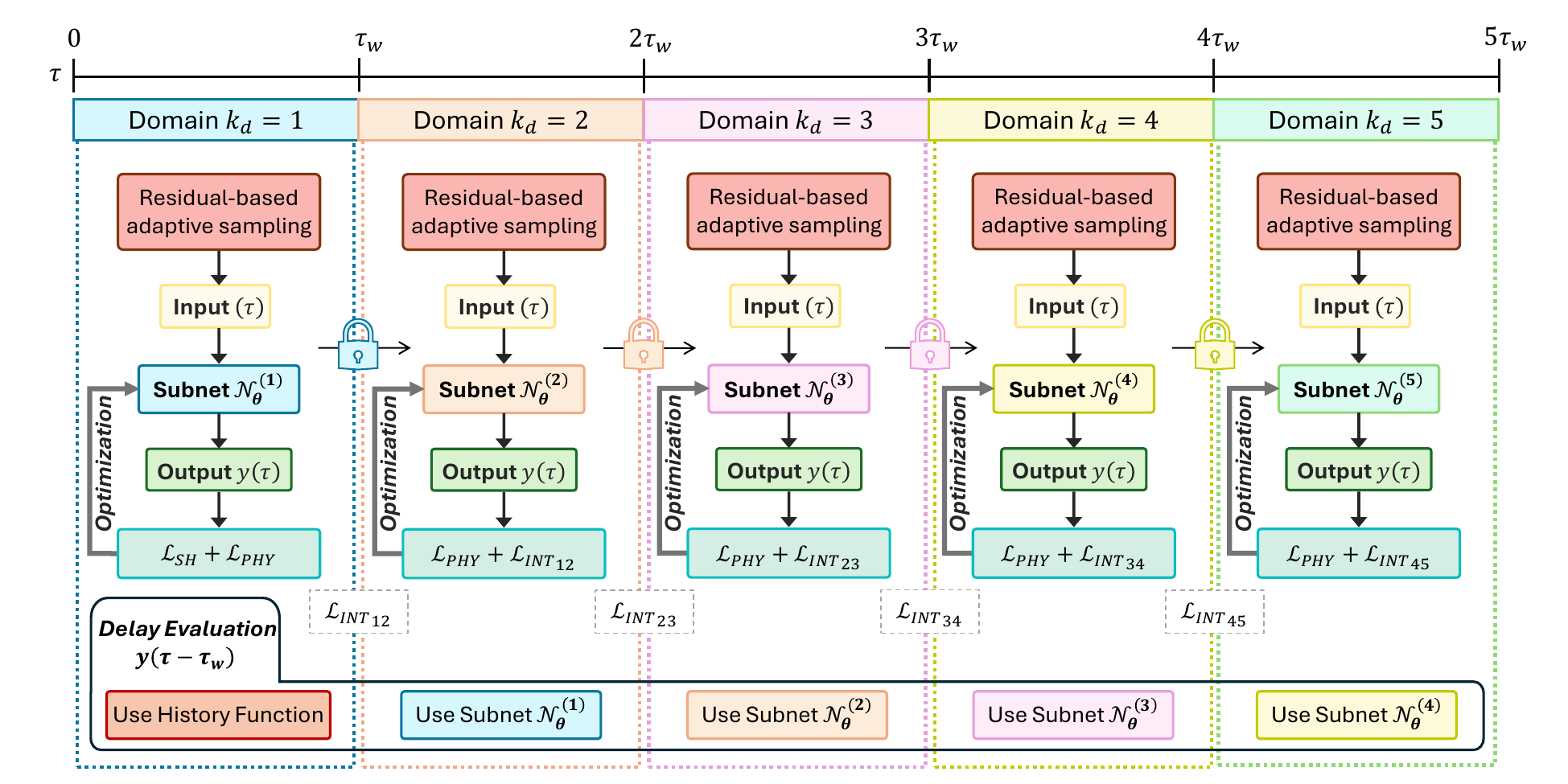}
        \caption{Schematic of the XPINN}
        \label{fig:sche-xpinn-for}
\end{figure*}

The XPINN formulation used to approximate the solution to the considered DDE is illustrated schematically in Fig.~\ref{fig:sche-xpinn-for}. The overall training length is partitioned into subdomains of each interval.
For the $k_d$-th domain, the training procedure proceeds through the following steps:
\begin{enumerate}
\item Freeze all previously trained subnetwork(s).
\item Activate only the current subnet for training.
\item Sample collocation points adaptively within the current interval.
\item Evaluate the residual, history, and interface constraints.
\item Compute the total loss and update the trainable weights using the Adam optimiser.
\item Repeat until the prescribed number of epochs is reached.
\end{enumerate}
\par Each of these components is described in detail in the following subsections. Together, they define the full workflow through which the XPINN framework enforces the governing DDE, maintains consistency across subdomains, and progressively refines the solution. 

\subsection{Data preparation and adaptive sampling}

\par
\begin{figure*}[!hbt]
        \centering
        \includegraphics[width=\linewidth]{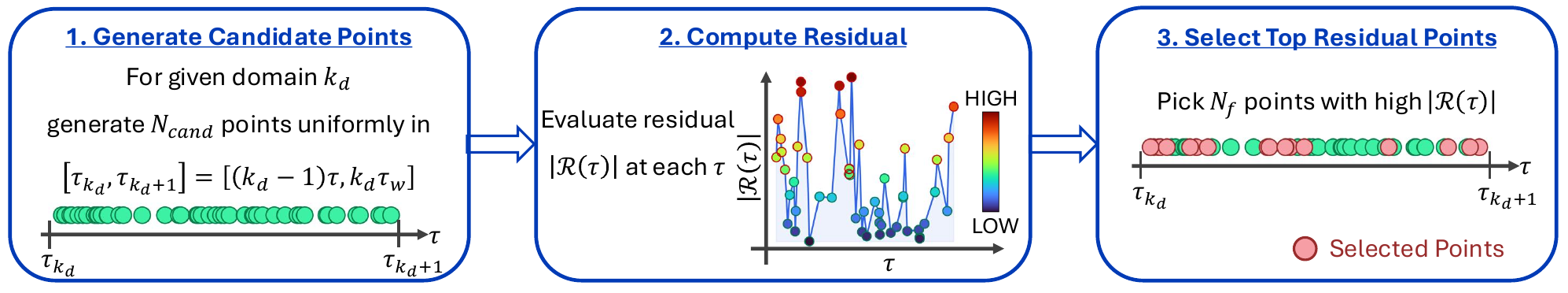}
        \caption {Schematic of residual-based adaptive sampling in XPINN at one epoch}
        \label{fig:xpinn-sampling}
\end{figure*}

 The overall time interval $\tau \in[0, 5\tau_w]$ is partitioned into multiple consecutive subdomains of equal length, each defined as $[(k_d-1)\tau_w, k_d\tau_w]$, for subdomain index $k_d$, $k_d \in \{1,\dots, 5\}$. Also, a separate ANN subnet $\mathcal{N}_\theta^{(k_d)}$ is assigned to each subinterval. To improve the efficiency of collocation-based training, the adaptive residual sampling procedure is implemented. Fig.~\ref{fig:xpinn-sampling} shows the three steps involved in the residual-based sampling procedure. First, for a given subdomain $k_d$, a set of $N_{cand}$ candidate points is uniformly generated within the corresponding interval. For each candidate point, the local residual of the XPINN model corresponding to the domain $k_d$ is evaluated by enabling the AD and calculating the absolute value of the DDE residual using Eq.~\eqref{Eq:residualDDE}. The resulting residual field is then used to identify the data points with the largest approximation errors. Finally, the $N_f$ points associated with the highest residual values are selected as the collocation points for that domain. 
This adaptive sampling procedure is repeated at every epoch to ensure that the collocation points continually track the regions where the model exhibits the largest residuals. 
\par To improve ANN training, the physical time interval $[\tau_{start}, \tau_{end}]$ was normalized to the interval $[-1,1]$ using the affine transformation
\begin{equation}
s(\tau) = 2\frac{\tau-\tau_{start}}{\tau_{end}-\tau_{start}} - 1,
\end{equation}
with inverse mapping
\begin{equation}
\tau(s) = \frac{1}{2}(s+1)(\tau_{end}-\tau_{start}) + \tau_{start}.
\end{equation}
This normalisation maps the input to a bounded, standardised domain, which typically improves optimisation performance and mitigates scaling issues during training. 
\subsection{XPINN subnet architecture}

\par
\begin{figure*}[!hbt]
        \centering
        \includegraphics[width=\linewidth]{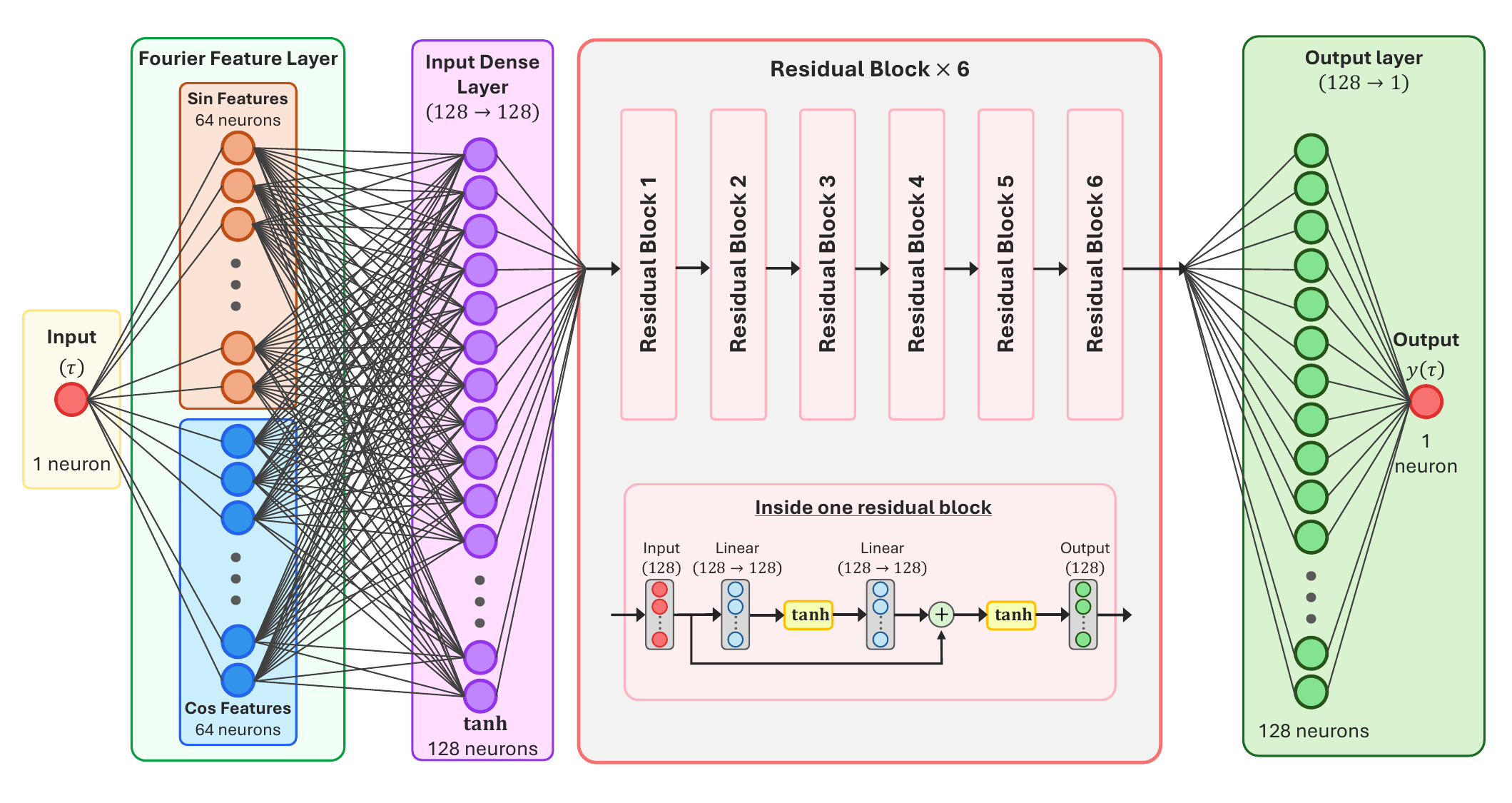}
        \caption{Network architecture of the subnet in XPINN}
        \label{fig:sche-xpinn-subnet}
\end{figure*}

 This section details the architecture and the operational flow of the subnet within the considered XPINN framework. The schematic representation of a subnet is depicted in Fig.~\ref{fig:sche-xpinn-subnet}.
For a given subdomain $k_d$, the normalised collocation points are first passed into the subnet stage. To better capture how the solution changes with time, the time input is first transformed using a Fourier feature embedding. Specifically, for a normalised input time variable $s$, the embedding maps it into a higher-dimensional sinusoidal feature space of the form
\begin{equation}
\mathcal{F}(s) = \left[\sin(2\pi \mathbf{B}_{k_d}s), \cos(2\pi \mathbf{B}_{k_d}s)\right],
\end{equation}

where $\mathbf{B}_{k_d}\in\mathbb{R}^{64\times1}$ is a non-time varying random projection matrix. The entries of $\mathbf{B}_{k_d}$ are independently sampled from a zero-mean Gaussian distribution with standard deviation $5$. The resulting 64 sine and 64 cosine components provide a 128-dimensional feature representation of the temporal coordinate. The projection matrix is initialised once for each subnetwork and kept fixed throughout optimisation, so it is not a trainable parameter. The resulting mapping provides a set of sinusoidal features spanning a range of characteristic frequencies. This input encoding is particularly useful for the present delayed machining dynamics, where the solution may exhibit multiple temporal scales associated with structural vibration, regenerative feedback, and frictional effects.

After the embedding layer, the encoded input is passed through a linear input layer with $\tanh$ activation and subsequently propagated through a sequence of residual blocks. Each residual block applies two linear layers with a $\tanh$ activation, together with a skip connection that preserves the incoming state. For a hidden state vector $\mathbf{Z}$, a typical residual block can be written as

\begin{align}
\mathbf{Z_r} &= \mathbf{W}_2 \tanh(\mathbf{W}_1 \mathbf{Z} + \mathbf{b}_1) + \mathbf{b}_2, \\
\mathbf{Z}_{\text{out}} &= \tanh(\mathbf{Z} + \mathbf{Z_r}),
\end{align}
where $\mathbf{W}_1$, $\mathbf{W}_2$ are trainable weight matrices and $\mathbf{b}_1$, $\mathbf{b}_2$ are bias vectors. The residual connection improves gradient flow and helps stabilise training for ANNs. The full XPINN model consists of five subnetworks, denoted by
\begin{equation}
\mathcal{N}_\theta^{(k_d)}, \qquad k_d \in \{1,\dots, 5\},
\end{equation}
each defined on a local temporal interval. 
\subsection{Delay evaluation and total loss}
The delayed state $(\tau-\tau_w)$ is evaluated differently depending on the current subdomain. If the delayed time $\tau-\tau_w$ lies within the prescribed history interval, the history function is used. Otherwise, the delayed value is obtained from the subnet associated with the preceding subdomain. This causal construction ensures that the solution at the current time depends only on already available past information. In other words, for subdomain 1, the subnet $\mathcal{N}_\theta^{(1)}$ uses the history function provided, while for subsequent subdomains $k_d=2,3,4,5$, the subnets $\mathcal{N}_\theta^{(2)}, \mathcal{N}_\theta^{(3)}, \mathcal{N}_\theta^{(4)}, \mathcal{N}_\theta^{(5)},$ respectively use the previously trained subnets $\mathcal{N}_\theta^{(1)}, \mathcal{N}_\theta^{(2)}, \mathcal{N}_\theta^{(3)}, \mathcal{N}_\theta^{(4)}$.

\par The total loss for the XPINN is given by,
\begin{subequations}\label{eq:loss_all}
    \begin{equation}\label{Eq:loss_dm_1}
        \mathcal{L}_{Total}^{(k_d)} = \lambda_{PHY} \mathcal{L}_{PHY} + \lambda_{SH} \mathcal{L}_{SH}, \qquad \mathrm{when\;}k_d=1, 
    \end{equation}
    \begin{equation}\label{Eq:loss_dm_others}
        \mathcal{L}_{Total}^{(k_d)} = \lambda_{PHY} \mathcal{L}_{PHY} + \lambda_{INT} \mathcal{L}_{INT}^{(k_d-1, k_d)}, \qquad \mathrm{when\;}k_d\in[2,5], 
    \end{equation}
\end{subequations}
where the physics-informed loss $\mathcal{L}_{PHY}$ is the same for all the domains, and it is the same as the SPINN formulation given in Eq.~\eqref{Eq:phy_loss}. For the first subdomain, the total loss given in Eq.~\eqref{Eq:loss_dm_1} is used. $\mathcal{L}_{SH}$ in Eq.~\eqref{Eq:loss_dm_1} represents the shifted history anchoring loss, and it is calculated as
\begin{equation}
    \mathcal{L}_{SH} = \frac{1}{N_H}\sum_{i=1}^{N_H}\left|\mathcal{N}_\theta^{(1)}\left(\tau_i^{(H)}\right)-\Phi_H\left(\tau_i^{(H)}-\tau_w\right)\right|^2, 
\end{equation}
where $N_H$ is the number of points sampled over the first temporal subdomain: $\tau_i^{(H)}\in[0,\tau_w]$. The shifted-history anchoring loss $\mathcal{L}_{SH}$ is introduced to provide a dense training constraint for the first temporal subdomain. Unlike a pointwise initial-condition loss, which constrains the network only at $\tau=0$, $\mathcal{L}_{SH}$ constrains the first subnet $\mathcal{N}_\theta^{(1)}$ at multiple points throughout $[0,\tau]$. This improves the initialisation of the first subdomain and reduces the propagation of approximation errors to subsequent temporal subdomains. The other subdomains use the loss function provided in Eq.~\eqref{Eq:loss_dm_others}. The second term $\mathcal{L}_{INT}^{(k_d-1, k_d)}$ in Eq.~\eqref{Eq:loss_dm_others} is introduced to ensure smooth propagation of the solution from one domain to the next by enforcing the interface continuity at each internal boundary as given below,
\begin{equation}
    \mathcal{L}_{INT}^{(k_d-1, k_d)}= \left[y_\theta^{(k_d-1)}(\tau_{int})-y_\theta^{(k_d)}(\tau_{int})\right] + \left[\dot{y}_\theta^{(k_d-1)}(\tau_{int})-\dot{y}_\theta^{(k_d)}(\tau_{int})\right],
\end{equation}
where $\tau_{int}$ denotes the interface point between the subdomains $(k_d-1)$ and $k_d$. The quantity $y_\theta^{(k_d)}$ represents the solution produced the subnet $\mathcal{N}_\theta^{(k_d)}$, while $\dot{y}_\theta^{(k_d)}$ denotes its time derivate calculated via AD. 
\par $\lambda_{PHY}$, $\lambda_{SH}$, and $\lambda_{INT}$ in Eq.~\ref{eq:loss_all} denote the adaptive weights of the loss function, and their values are considered as
\begin{subequations}
    \begin{equation}
        \lambda_{PHY}=1,
    \end{equation}
    \begin{equation}
    \lambda_{SH}=\frac{\operatorname{SG}\!\left(\mathcal{L}_{PHY}\right)+\Gamma}{\operatorname{SG}\!\left(\mathcal{L}_{SH}\right)+\Gamma},
    \end{equation}
    \begin{equation}
    \lambda_{INT}=\frac{\operatorname{SG}\!\left(\mathcal{L}_{PHY}\right)+\Gamma}{\operatorname{SG}\!\left(\mathcal{L}_{INT}\right)+\Gamma},
    \end{equation}
\end{subequations}
where $\Gamma$ is a small positive regularisation constant introduced to avoid division by zero, and $\operatorname{SG}(\cdot)$ denotes the stop-gradient operation, which evaluates the numerical value of its argument while preventing gradients from being propagated through it during backpropagation. In the current implementation, the stop-gradient operation is implemented using PyTorch's \texttt{detach()} function. The value of $\Gamma$ is set to $10^{-12}$. This adaptive weighting strategy balances the contributions of the different constraints during training and helps stabilise the optimisation process across the subdomains.
\subsection{Training Implementation}
\par Training is performed sequentially across the subdomains. The first subnet is trained using the history constraint and physics loss on the initial interval. Once its parameters are optimised, the network is frozen and no longer updated. The second subnet is then trained using the frozen output of the first subnet to evaluate the delayed state, together with interface continuity constraints. This process is repeated for all remaining subdomains.

The optimisation uses the Adam algorithm with an exponentially decaying learning rate. The learning rate $\eta_{epo}$ at epoch $epo$ is updated according to
\begin{equation}
\eta_{epo} = \gamma_d \eta_{epo-1},
\end{equation}
where the initial learning rate is set to $\eta_0=10^{-3}$, and $\gamma_d=0.99995$ is the exponential decay factor. The learning rate is therefore reduced incrementally at each epoch, resulting in a cumulative reduction of approximately $39.35\%$ after $10^4$ epochs and helping the optimiser converge more smoothly.

The resulting training strategy is domain-decomposed and physics-constrained. It ensures that information flows forward in time only through previously trained subnetworks, which is essential for solving DDEs with regenerative dependence. For the $k_d$-th subdomain, the model parameters of subnetworks $1,\dots,k_d-1$ are set to non-trainable, while the parameters of only the $k_d$-th network are kept trainable. The optimiser is then built using only the parameters that currently require gradients. At each epoch, the collocation points are resampled, the physics residual is evaluated, and the total loss is minimised using backpropagation. This training strategy is particularly well-suited for the long-term integration of DDEs because it avoids the instability that can arise when a single global network is forced to represent the entire history across multiple delay intervals.
\begin{figure}[!hbt]
    \centering
    \includegraphics[width=\linewidth]{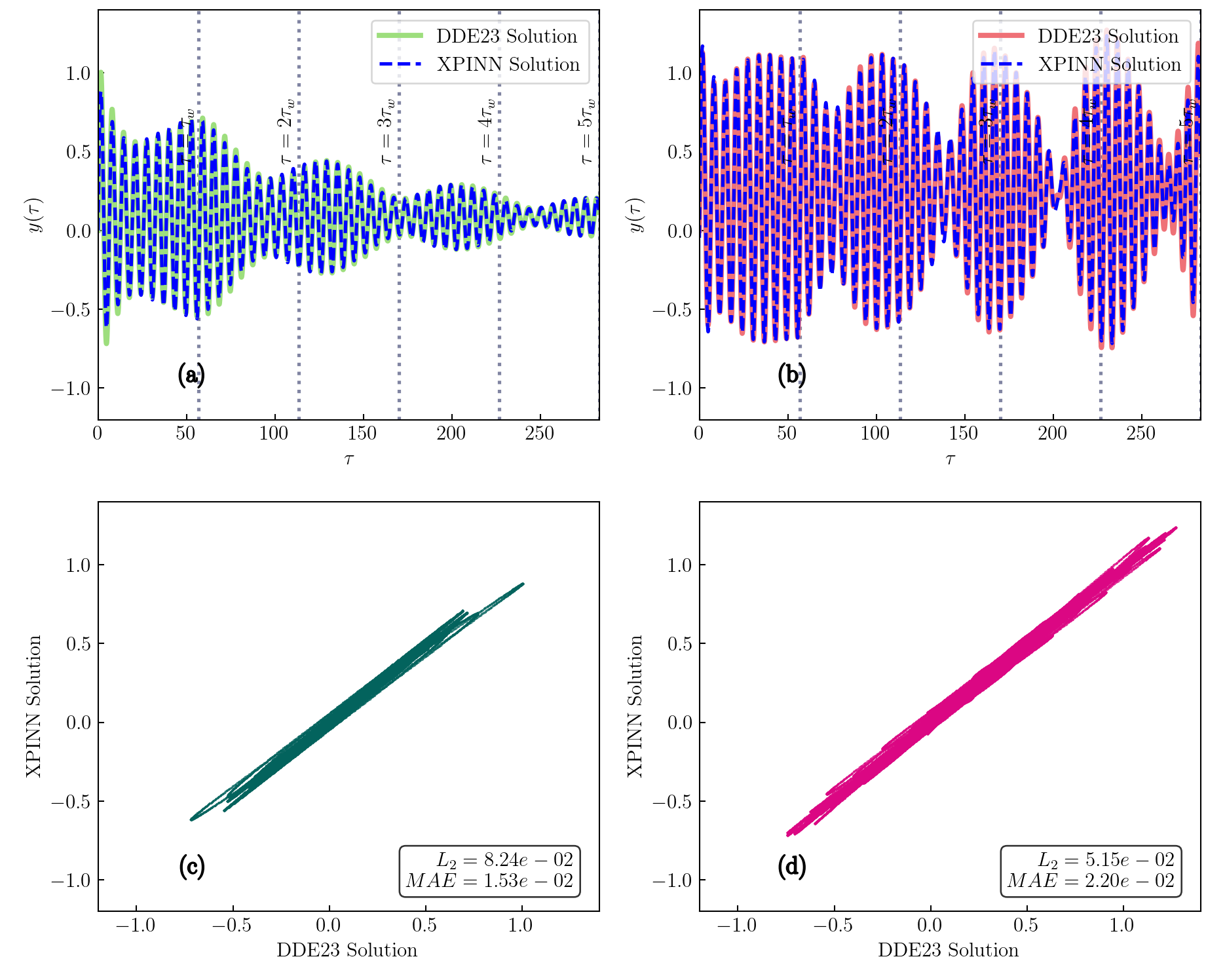}
    \caption{XPINN results in solving the machining system for simulation time $\tau\in[0, 5\tau_w]$ with a non-constant history function $\Phi_H(\tau)=\sin(\tau)$. (a) and (b) represent plots of predicted values (dotted line) over the DDE23 reference solution (solid line) for the stable and unstable cases, respectively. (c) and (d) correspond to the scatter plots of the DDE solution with the PINN solution, respectively, for the stable and unstable cases.}
    \label{fig:xpinn-sin-for}
\end{figure}
\subsection{Data-driven solution of machining system using XPINNs}

\noindent With all methodological components established, the next step is to assess the capability of the XPINN framework to approximate solutions over long simulation horizons, particularly in the presence of non-constant history functions. To this end, the sampling parameters are fixed at $N_{cand}=7.5\times10^4$ and $N_f=5\times10^4$. The system parameters are kept identical to those used in the SPINN formulation for both stable and unstable configurations, ensuring a consistent basis for comparison between the two approaches. As mentioned earlier, the overall time interval $\tau\in[0, 5\tau_w]$ is divided into 5 subdomains, and the training is carried out for $8\times10^4$ epochs for each domain. Fig.~\ref{fig:xpinn-sin-for} shows the comparison between the XPINN approximated solution and the DDE23 solution. From Fig.~\ref{fig:xpinn-sin-for}~(a)\&(b), it is evident that the XPINN solution, shown as dotted lines, closely overlaps with the DDE23 reference solution, plotted as solid lines. This agreement demonstrates that XPINNs can accurately approximate the solution of the considered machining DDE in both stable and unstable regimes. The straight band of scattered points in Fig.~\ref{fig:xpinn-sin-for}~(c)\&(d) further corroborates the outcomes. This visual agreement is supported quantitatively by the error metrics. The computed relative $L_2$ error is $8.24\times10^{-2}$ and $5.15\times10^{-2}$, respectively, for stable and unstable cases, confirming that the global deviation between the XPINN and DDE23 solutions remains small across all the subdomains. Likewise, the MAE values of $1.53\times10^{-2}$ and $2.2\times10^{-2}$ indicate that the point-wise discrepancies are uniformly bounded and do not accumulate over the long simulation. Together, these visual and numerical results validate the robustness of the proposed XPINN formulation in capturing the system's long-term dynamics under non-constant history functions.

\section{Inverse PINN framework for estimating the machining parameters}\label{sec5}
\noindent The forward problem presented in Section \ref{sec4} assumes that the governing system parameters of the machining system are known and determines the corresponding dynamic response. In practical machining applications, however, several parameters are difficult to determine directly from experiments. In particular, the cutting force coefficient ($K$) and the process damping coefficient ($C_y$) strongly influence the dynamic response. The values of the parameters considered substantially large and vary in magnitude with $K=6.02\times10^{9}\; [{\rm Nm^{-2}}]$ and $C_y=6.11\times10^{5}\;[{\rm Nm^{-1}}]$. Directly treating these as trainable ANN quantities can therefore introduce an unfavourable scale disparity between the optimisation variables and may affect the numerical stability of the inverse training process.

\par To improve the conditioning of the optimisation, the logarithms of the unknown parameters are used as the trainable variables. $K_{\log}=\log(K)$ and $C_{y,\log}=\log(C_y)$ are introduced and the values are now become $K_{\log}\approx 22.5183531 $ and $C_{y,\log} \approx 13.3228524$. This exponential transformation provides two advantages for the inverse optimisation. (i) It ensures that the estimated parameters remain strictly positive throughout training, which is consistent with the physical meaning of the unknown coefficients. (ii) Optimisation is performed on a logarithmic scale, which reduces the disparity in magnitude between the two parameters and allows their relative changes to be learned more effectively. The transformed parameters are used directly within the non-dimensional model. Thus, wherever they appear in the governing DDE, they are evaluated as
\begin{equation}
    K=\exp(K_{\log}),\qquad C_y=\exp(C_{y,\log}).
\end{equation}
For example, the dimensionless cutting-force parameter and process-damping coefficient become
\begin{equation}
    W=a_p\frac{\exp(K_{\log})}{k}, \qquad c_y=\frac{30\exp(C_{y,\log})}{\pi R\exp(K_{\log})}.
\end{equation}
\par The considered unknown parameters are treated as trainable quantities,
\begin{equation}
    \begin{bmatrix}
K^{\mathrm{train}}_{\log} & C^{\mathrm{train}}_{y,\log}
\end{bmatrix}^{T},
\end{equation}
and are optimised together with the ANN's weights and biases. The ANN output $y_\theta(\tau)$ is substituted into the governing DDE together with the current estimates of $K_{\log}$ and $C_{y,\log}$. This modifies the physics residual $\mathcal{R}(\tau;\theta)$ given in Eq.~\eqref{Eq:residualDDE} to $\mathcal{R}(\tau;\theta,K_{\log}, C_{y,\log})$. 
\begin{figure*}[!hbt]
    \centering
    \includegraphics[width=.9\linewidth]{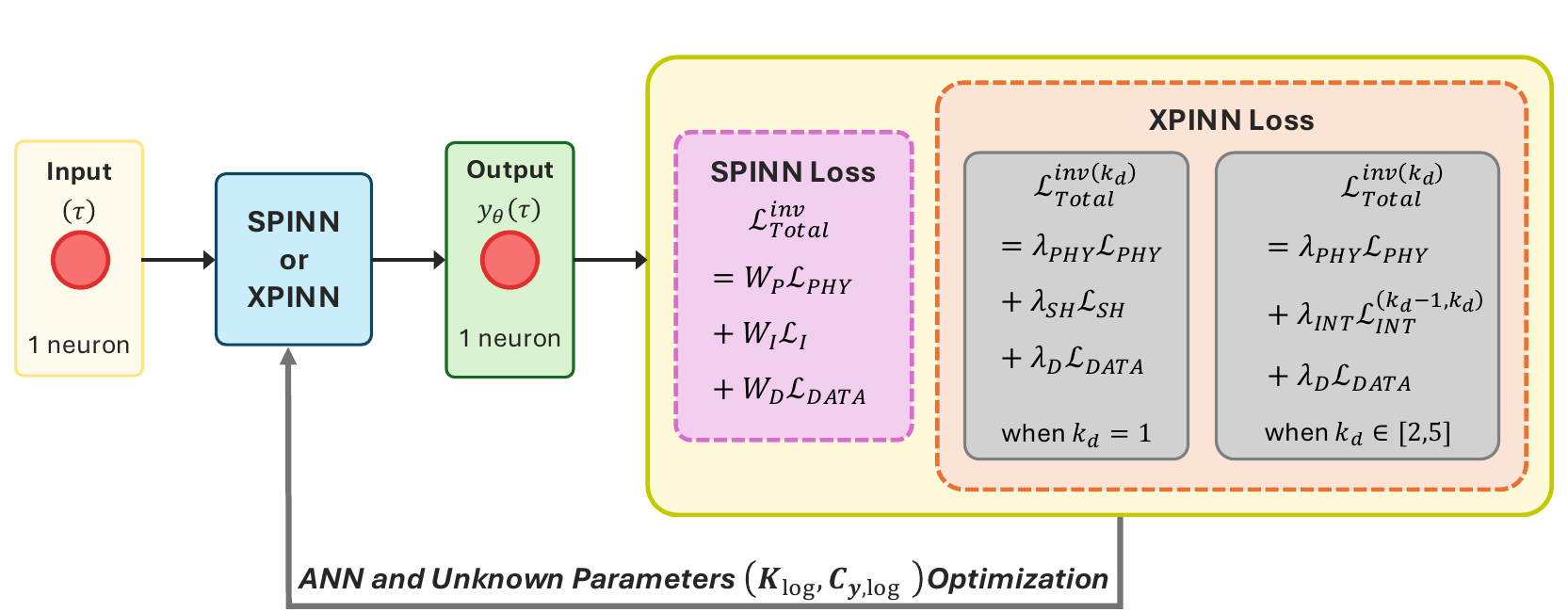}
    \caption{Schematic of the inverse PINN for estimating the unknown parameters of the considered machining system.}
    \label{fig:sche-Ipinn}
\end{figure*}
\par Fig.~\ref{fig:sche-Ipinn} shows the schematic of the general inverse framework used in this work. In an inverse PINN, the unknown physical parameters are introduced as additional trainable variables and are optimised simultaneously with the neural-network parameters. Consequently, the network is not only required to approximate the system response but also to identify the parameter values that best match the governing DDE and the measured response. In addition to the losses that are already there from the forward framework, the data loss is also included in the total loss. It ensures that the PINN solution is not only consistent with the governing DDE but also agrees with the measured machining response. This modifies the total loss as presented in Eq. \ref{Eq:loss_spinn_inverse}  and Eq. \ref{eq:loss_all_inverse}. The total loss for the inverse SPINNs (I-SPINN) is represented as
\begin{subequations}
   
    \begin{equation}\label{Eq:loss_spinn_inverse}
        \mathcal{L}_{Total}^{inv} = W_P \mathcal{L}_{PHY} + W_I \mathcal{L}_{I} + W_D \mathcal{L}_{DATA},
    \end{equation}
with
\begin{equation}
    W_{P} = \dfrac{\mathcal{L}_{PHY}}{\mathcal{L}_{PHY}+\mathcal{L}_I + \mathcal{L}_{DATA}}, 
\end{equation}
\begin{equation}
    W_I =  \dfrac{\mathcal{L}_{I}}{\mathcal{L}_{PHY}+\mathcal{L}_I + \mathcal{L}_{DATA}},
\end{equation}
\begin{equation}
     W_D =  \dfrac{\mathcal{L}_{DATA}}{\mathcal{L}_{PHY}+\mathcal{L}_I + \mathcal{L}_{DATA}},
\end{equation}
\end{subequations}

and, for Inverse XPINNs (I-XPINN)
\begin{subequations}\label{eq:loss_all_inverse}
    \begin{equation}\label{Eq:loss_dm_1_inverse}
        \mathcal{L}_{Total}^{inv(k_d)} = \lambda_{PHY} \mathcal{L}_{PHY} + \lambda_{SH} \mathcal{L}_{SH}+\lambda_{D} \mathcal{L}_{DATA}, \qquad \mathrm{when\;}k_d=1, 
    \end{equation}
    \begin{equation}\label{Eq:loss_dm_others_inverse}
        \mathcal{L}_{Total}^{inv(k_d)} = \lambda_{PHY} \mathcal{L}_{PHY} + \lambda_{INT} \mathcal{L}_{INT}^{(k_d-1, k_d)}+\lambda_{D} \mathcal{L}_{DATA}, \qquad \mathrm{when\;}k_d\in[2,5], 
    \end{equation}

    \end{subequations}
    with
    \begin{subequations}
    \begin{equation}
        \lambda_{PHY}=1,
    \end{equation}
    \begin{equation}
    \lambda_{SH}=\frac{\operatorname{SG}\!\left(\mathcal{L}_{PHY}\right)+\Gamma}{\operatorname{SG}\!\left(\mathcal{L}_{SH}\right)+\Gamma},
    \end{equation}
    \begin{equation}
    \lambda_{INT}=\frac{\operatorname{SG}\!\left(\mathcal{L}_{PHY}\right)+\Gamma}{\operatorname{SG}\!\left(\mathcal{L}_{INT}\right)+\Gamma},
    \end{equation}
    \begin{equation}
    \lambda_{D}=\frac{\operatorname{SG}\!\left(\mathcal{L}_{PHY}\right)+\Gamma}{\operatorname{SG}\!\left(\mathcal{L}_{DATA}\right)+\Gamma},
    \end{equation}
\end{subequations}
where,
\begin{equation}\label{eq-data_loss}
     \mathcal{L}_{DATA} = \dfrac{1}{N_{data}}\sum_{i=1}^{N_{data}}\left|y_\theta-y_{ref}\right|^2
\end{equation}
represents the data loss and measures the root mean square error between the response predicted by the PINN framework and the available reference or observation data. $W_D$ in Eq.~\eqref{Eq:loss_spinn_inverse} and $\lambda_{D}$ in Eq.~\eqref{eq:loss_all_inverse} represent the adaptive weights for the case of I-SPINN and I-XPINN, respectively. After training, the estimated parameters are quantitatively evaluated by calculating the relative percentage error between the true parameter values, $\{K, C_y\}$, and the PINN-estimated parameters obtained by transforming the predicted logarithmic values $\{K_{\log}, C_{y,\log}\}$ back to their physical scale using the exponential function. Specifically,
\begin{equation}
    K_{\mathrm{pred}} = \exp{(K_{\log})}, \qquad C_{y, \mathrm{pred}} = \exp{(C_{y, \log})},
\end{equation}
and the relative percentage errors are calculated as

\begin{equation}\label{eq-percent-error}
    E_K = \dfrac{\left|K_{\mathrm{pred}}-K_{\mathrm{true}}\right|}{\left|K_{\mathrm{true}}\right|}\times 100,
\qquad
    E_{C_y} = \dfrac{\left|C_{y,\mathrm{pred}}-C_{y,\mathrm{true}}\right|}{\left|C_{y,\mathrm{true}}\right|}\times 100.
\end{equation}

\par Here $K_{\mathrm{true}}$ and $C_{y,\mathrm{true}}$ denote the true reference parameter values and have the values of $6.02\times10^{9}$ and $6.11\times10^{5}$, respectively.

\begin{figure*}[!hbt]
    \centering
    \includegraphics[width=\linewidth]{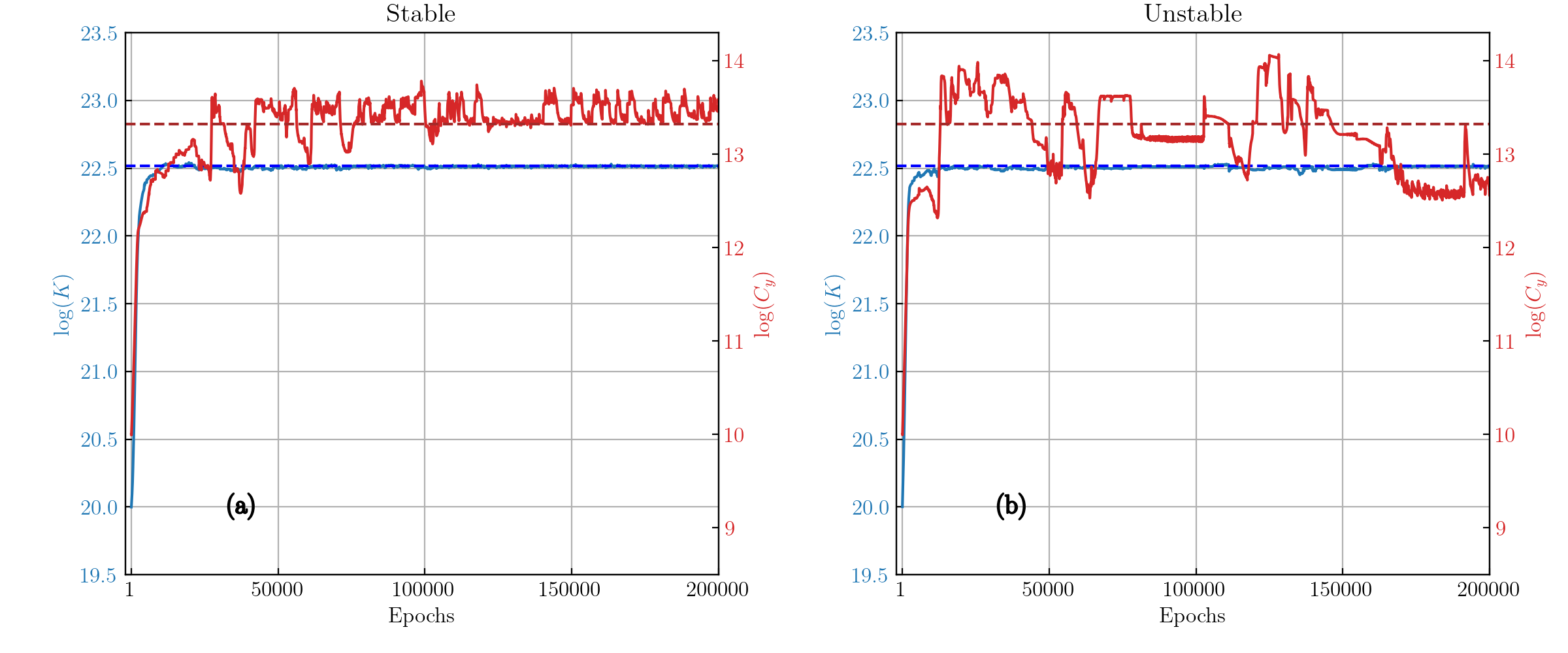}
    \caption{I-SPINN results in estimating the values of $K_{\log}$ and $C_{y,\log}$ of machining system \eqref{eq-nondim} for simulation time $\tau\in[0, 1.3\tau_w]$ with a constant history function $\Phi_H(\tau)=1$. (a) and (b) represent the values of $K_{\log}$ and $C_{y,\log}$ over the epochs for stable and unstable cases respectively. The blue and red dotted horizontal lines indicate the true values of $K_{\log}$ and $C_{y,\log}$, respectively.}
    \label{fig:spinn-inv-const}
\end{figure*}
\subsection{Parameter estimation using I-SPINNs}
\noindent The inverse SPINN formulation is first employed to examine whether it can reliably recover the cutting‑force and process‑damping coefficients for the given reference data. In this case, the time interval is set to $\tau\in[0, 1.3\tau_w]$ and the history function is taken as $\Phi_H(\tau)=1$, similar to that of the forward SPINN analysis. The initial guesses for the unknown parameters are set to $K_{\log}=20$ and $C_{y,\log}=10$. The I-SPINN framework is trained for $2\times10^5$ epochs. Fig.~\ref{fig:spinn-inv-const} represents the evolution of the two trainable machining system parameters, $K_{\log}$ and $C_{y,\log}$, for the stable and unstable parameter regimes as defined in Sec. \ref{sec2pt4}. The $x$-axis denotes the training epochs, while the left $y$-axis (in blue) shows the evolution of $K_{\log}$ and the right $y$-axis (in red) displays the values of $C_{y,\log}$. In both stable and unstable cases, the parameters progressively adjust throughout the training process and try to converge toward their respective true values, indicated by the horizontal dashed reference lines. The evolution of the cutting force coefficient $K_{\log}$ remains steady around its true value, whereas the damping coefficient exhibits oscillatory behaviour, with noticeably larger variations in the unstable case compared to the stable case.
\begin{figure}[!hbt]
    \centering
    \includegraphics[width=\linewidth]{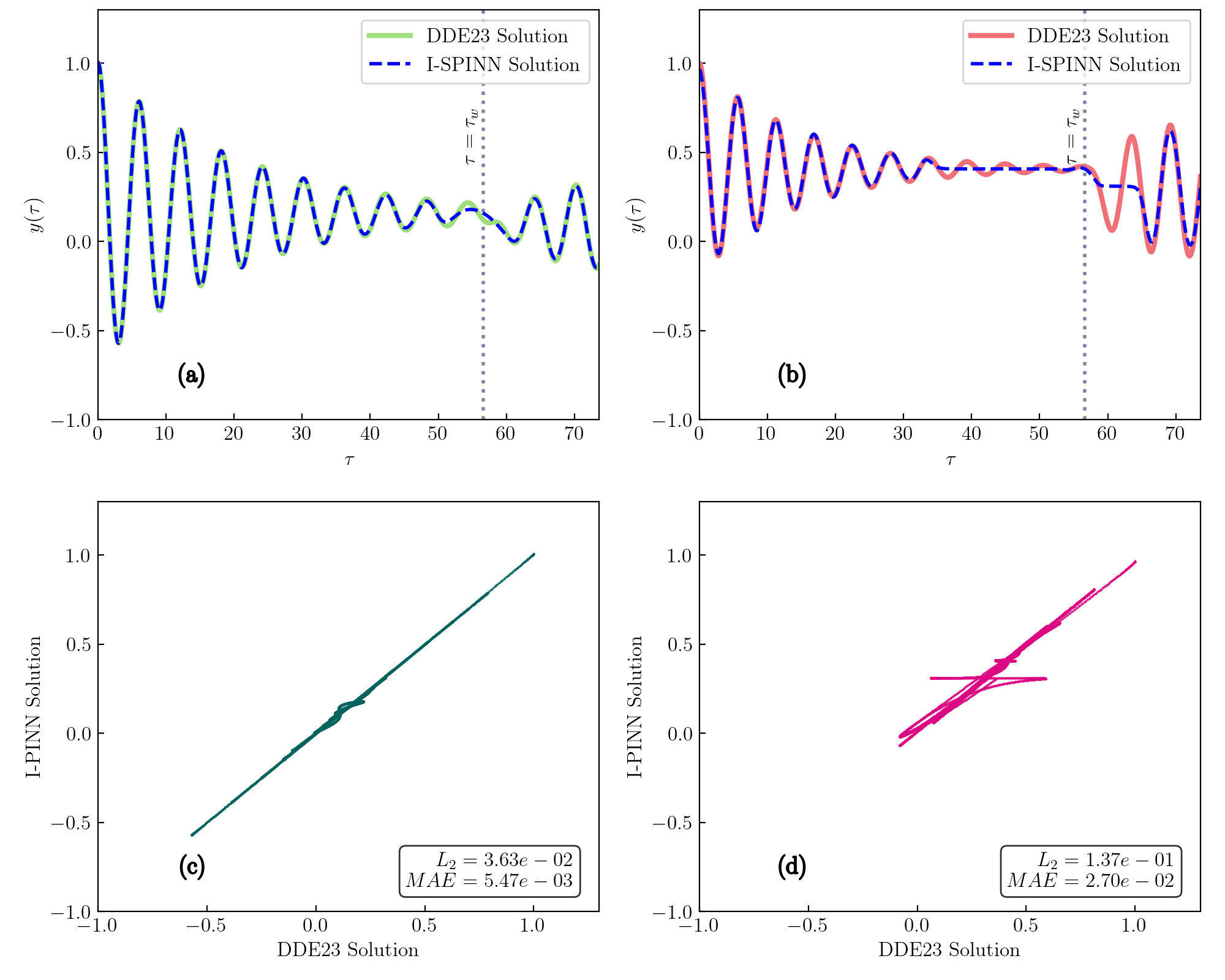}
    \caption{Comparison of the approximated solution with the reference solution of the considered DDE for (a) the stable case and (b) the unstable case with the I-SPINN framework. (c) and (d) correspond to the scatter plots of the DDE solution with the I-SPINN solution, respectively, for the stable and unstable cases.}
    \label{fig:spinn-inv-prediction}
\end{figure}
This observation is quantitatively verified by computing the relative error using Eq.~\eqref{eq-percent-error}. For the stable case, the estimated parameters yield errors of 
$E_K=0.6088\%$ and $E_{C_y} = 26.0348\%$, whereas for the unstable case, the corresponding errors are $E_K=1.0277\%$ and $E_{C_y} = 53.3091\%$. These results indicate that the I-SPINN fails to estimate the process damping coefficient $C_y$, with the estimation accuracy deteriorating for the unstable case. Nevertheless, the identified parameters are subsequently used to reconstruct the solution of the DDE over the considered interval $[0, 1.3\tau_w]$, and the results are depicted in Fig.~\ref{fig:spinn-inv-prediction}. It is clear from  Fig.~\ref{fig:spinn-inv-prediction} that the stable case exhibits substantially lower prediction errors with $L_2=3.63\times 10^{-2}$ and $MAE=5.47\times 10^{-3}$, whereas the unstable case exhibits considerably higher errors of $L_2=1.37\times 10^{-1}$ and $MAE=2.7\times 10^{-2}$. These results demonstrate that inaccuracies in the identified parameters, particularly $C_y$, can lead to substantial degradation in the reconstructed dynamics. The deterioration from the stable to the unstable case highlights the difficulty of I-SPINN in accurately estimating the governing parameters and subsequently reproducing the DDE solution in the unstable region, even for the given small time interval $\tau\in[0,1.3\tau_w]$ and a constant history function. The following analysis therefore examines the performance of the XPINN over an extended time horizon $\tau\in[0, 5\tau_w]$, with a non-constant history function to provide a more realistic and challenging problem to test the algorithm.

\subsection{Parameter estimation using I-XPINNs}
\par
\begin{figure*}[!hbt]
    \centering
    \includegraphics[width=\linewidth]{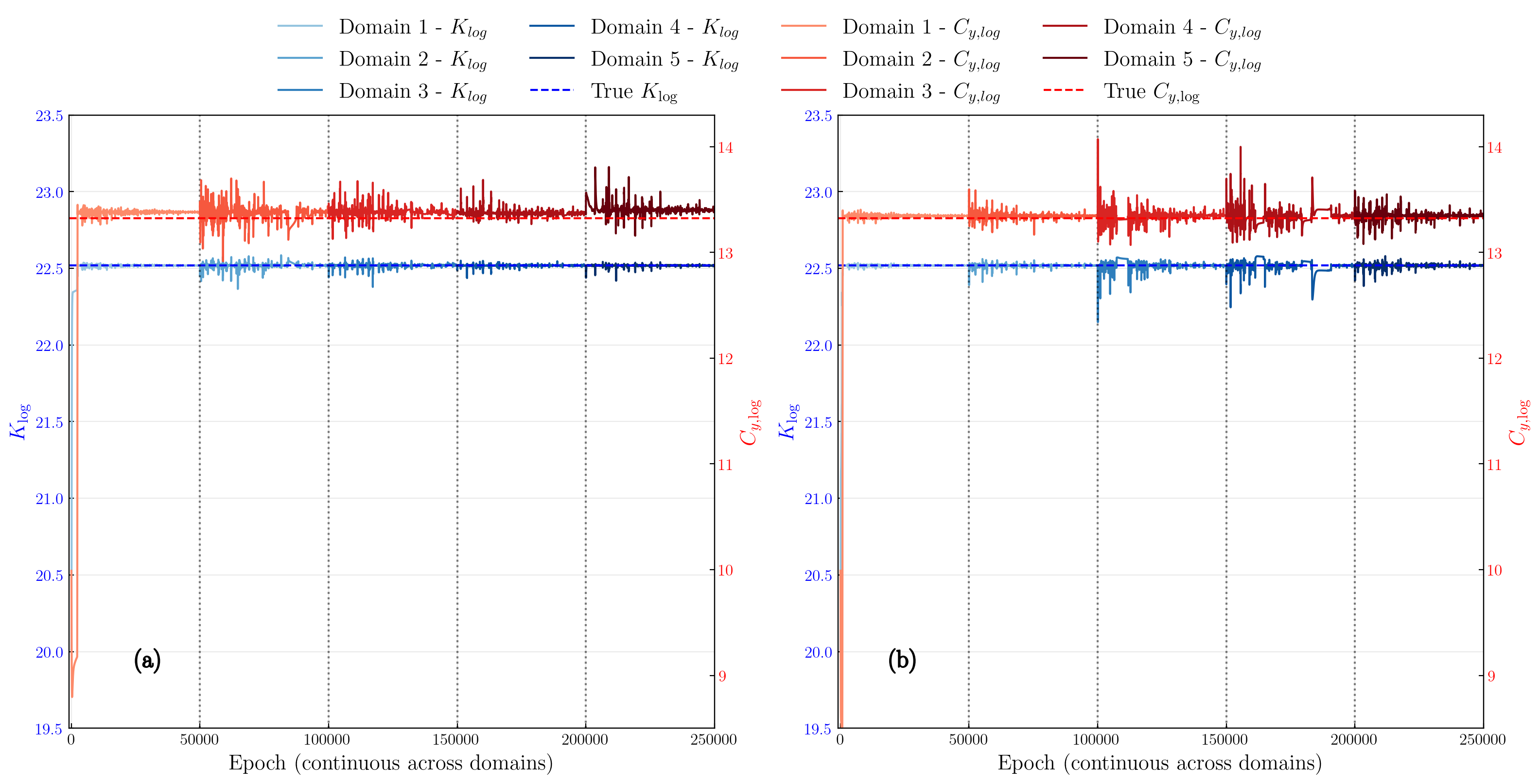}
    \caption{I-XPINN results in estimating the values of $K_{\log}$ and $C_{y,\log}$ for the (a) stable case and (b) unstable case reference data over the simulation interval $\tau\in[0, 5\tau_w]$ using the history function $\Phi_H(\tau)=\sin(\tau)$. The blue and red dotted horizontal lines indicate the true values of $K_{\log}$ and $C_{y,\log}$, respectively.}
    \label{fig:xpinn-inv-both}
\end{figure*}

 Similar to that of the forward formulation, the proposed I-XPINN formulation is now applied over the extended temporal domain $0\le\tau\le5\tau_w$. As discussed earlier, the extended domain is decomposed into multiple subdomains, and all the parameters are kept the same as in the forward XPINN framework. The learning rate of the trainable system parameters $\{K_{log}, C_{y,\log}\}$ is set to $10^{-2}$, and each domain is trained for $5\times10^4$ epochs, resulting in a total of $25\times10^4$ training epochs across all five domains. The sampling points are chosen as $N_f=5\times10^4$ with $N_{cand}$ of $7\times10^4$ points in each domain. The evolution of the estimated parameters during I-XPINN training is shown for both the stable and unstable cases in Figs.~\ref{fig:xpinn-inv-both}~(a) and (b), respectively. The horizontal blue and red lines denote the true values of $K_{\log}$ and $C_{y,\log}$, respectively, and the vertical dashed lines represent the boundaries between the adjacent subdomains. 
A notable feature from Figs.~\ref{fig:xpinn-inv-both}~(a)\&(b) is the presence of oscillations in the evolution of the parameter values at the beginning of each domain and particularly evident near the domain boundaries, but gradually diminishing as training progresses within each subdomain. This is due to the transition between consecutive temporal subdomains, where the solution in the newly activated domain must simultaneously satisfy the delayed dynamics, and the parameters remain consistent with the solution propagated from the preceding domain. The effect is more pronounced at the beginning of the domain because the newly introduced subnet initially has less information about the local solution and must adjust its parameters to satisfy the physics, data, and interface constraints. As training proceeds, these constraints progressively stabilise the unknown parameter estimates, which results in fewer oscillations and convergence toward the corresponding true reference values. 

\par The magnitude of these oscillations in the evolution of the unknown parameters is more pronounced for the unstable case shown in Fig.~\ref{fig:xpinn-inv-both}~(b), particularly in the estimation of $C_{y,\log}$. This indicates that the unstable dynamics amplify the sensitivity of parameter estimation to transitions between successive domains. 
\begin{table*}[!hbt]
\centering
\begin{tabular}{c c c c c}
\toprule
\multirow{2}{*}{Relative Error $\%$} & \multicolumn{2}{c}{Stable} & \multicolumn{2}{c}{Unstable} \\
\cmidrule(lr){2-3} \cmidrule(lr){4-5}
 & I-SPINN & I-XPINN & I-SPINN & I-XPINN \\
\midrule
$E_{K}$  & $0.6088\%$ & $0.0027\%$ & $1.0277\%$ & $0.0178\%$ \\
$E_{C_{Y}}$ & $26.0348\%$ & $7.4179\%$ &  $53.3091\%$ & $2.5507\%$ \\
\bottomrule
\end{tabular}
\caption{Comparison of relative errors of parameter estimations with I-SPINN and I-XPINN.}
\label{tab:error_compar_ispinn_iXPINN}
\end{table*}
\begin{figure}[!hbt]
    \centering
    \includegraphics[width=\linewidth]{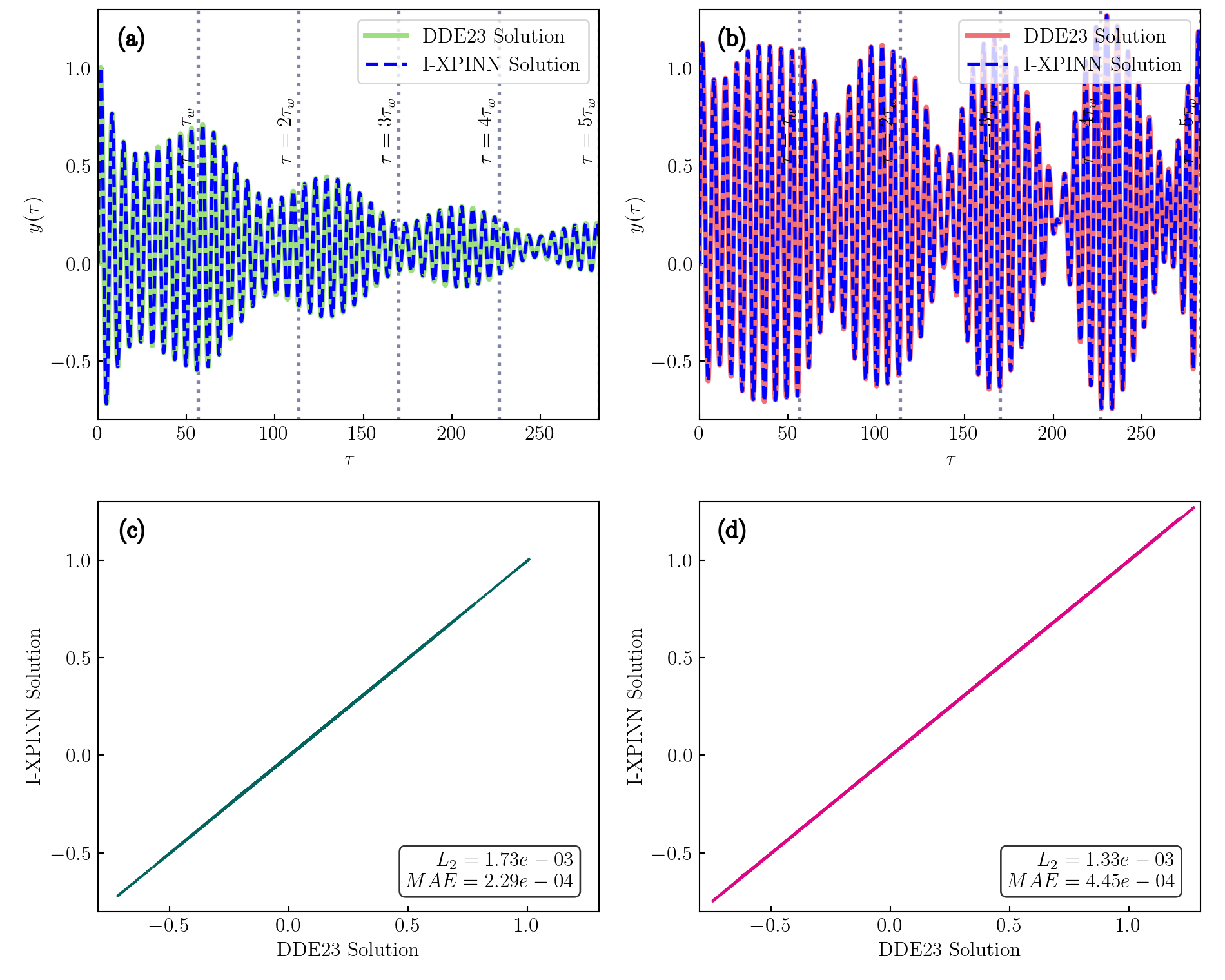}
    \caption{Comparison of the approximated solution with the reference solution of the considered DDE for (a) the stable case and (b) the unstable case with the I-XPINN framework. (c) and (d) correspond to the scatter plots of the DDE solution with the I-XPINN solution, respectively, for the stable and unstable cases.}
    \label{fig:xpinn-inv-prediction}
\end{figure}
The relative errors between the true and the final estimated parameter values are calculated using Eq.~\eqref{eq-percent-error}, and the error percentage values are compared with the outcomes of I-SPINN in Table~\ref{tab:error_compar_ispinn_iXPINN}. From Table~\ref{tab:error_compar_ispinn_iXPINN}, it is clear that the I-XPINN provides substantially improved parameter estimates. The relative error in $K$ is reduced to $0.0027\%$ for the stable case and $0.0178\%$ for the unstable case. More importantly, the relative error of $C_y$, $E_{C_y}$, decreases to $7.4179\%$ and $2.5507\%$, respectively, for the stable and unstable cases. Thus, the I-XPINN reduces the error in estimating $C_y$ by approximately $71.5\%$ in the stable case and $95.2\%$ in the unstable case compared with I-SPINN. Similarly, the error in $K$ is reduced by approximately $99.6\%$ for the stable case and $98.3\%$ for the unstable case. These results demonstrate a clear advantage of the I-XPINN formulation for inverse parameter estimation over the extended time domain. 

\par The I-XPINN framework is then used to reconstruct the solution of the considered DDE with the estimated parameters, and the results are presented in Fig.~\ref{fig:xpinn-inv-prediction}. As illustrated in Fig.~\ref{fig:xpinn-inv-prediction}, the I-XPINN prediction closely follows the reference solution over the entire extended time interval. The stable case achieves a relative $L_2$ error of $1.73\times 10^{-3}$ and an $MAE$ of $2.29\times 10^{-4}$, while the unstable case resulted in a comparable relative $L_2$ of $1.33\times 10^{-3}$ and an $MAE$ of $4.45\times 10^{-4}$. These error values obtained for both cases indicate that the I-XPINN maintains a high level of prediction accuracy regardless of whether the region is stable or unstable. Especially, the absence of a substantially high error value in the reconstruction of the DDE solution for the unstable case demonstrates the effectiveness of the domain-decomposition strategy and the I-XPINN framework in propagating the delayed dynamics across multiple subdomains while preserving accurate parameter estimation over the given extended time interval. All models were trained on an NVIDIA H100 NVL GPU. For SPINN, the training time was approximately 1.1 hours for the case of $N_f=5\times 10^4$ and approximately 5 hours when $N_f$ was increased to $25\times10^4$, while keeping 200,000 training epochs and the same $(5\tau_w)$ simulation horizon. The XPINN models required approximately 16.25 hours per case, with $N_{cand} = 7.5\times10^{4}$, $N_f=5\times10^4$ and $8\times10^4$ epochs per temporal subdomain. For the inverse parameter-identification problem, the total wall-clock training time was approximately 1.5 hours for SPINN and 10 hours for XPINN. The higher computational cost of XPINN is associated with its multi-domain architecture, separate subnetworks, interface constraints, adaptive residual-based sampling, and sequential domain-wise training.

\subsection{Parameter estimation using I-XPINNs for data with noise}
\par
\begin{figure*}[!hbt]
    \centering
    \includegraphics[width=\linewidth]{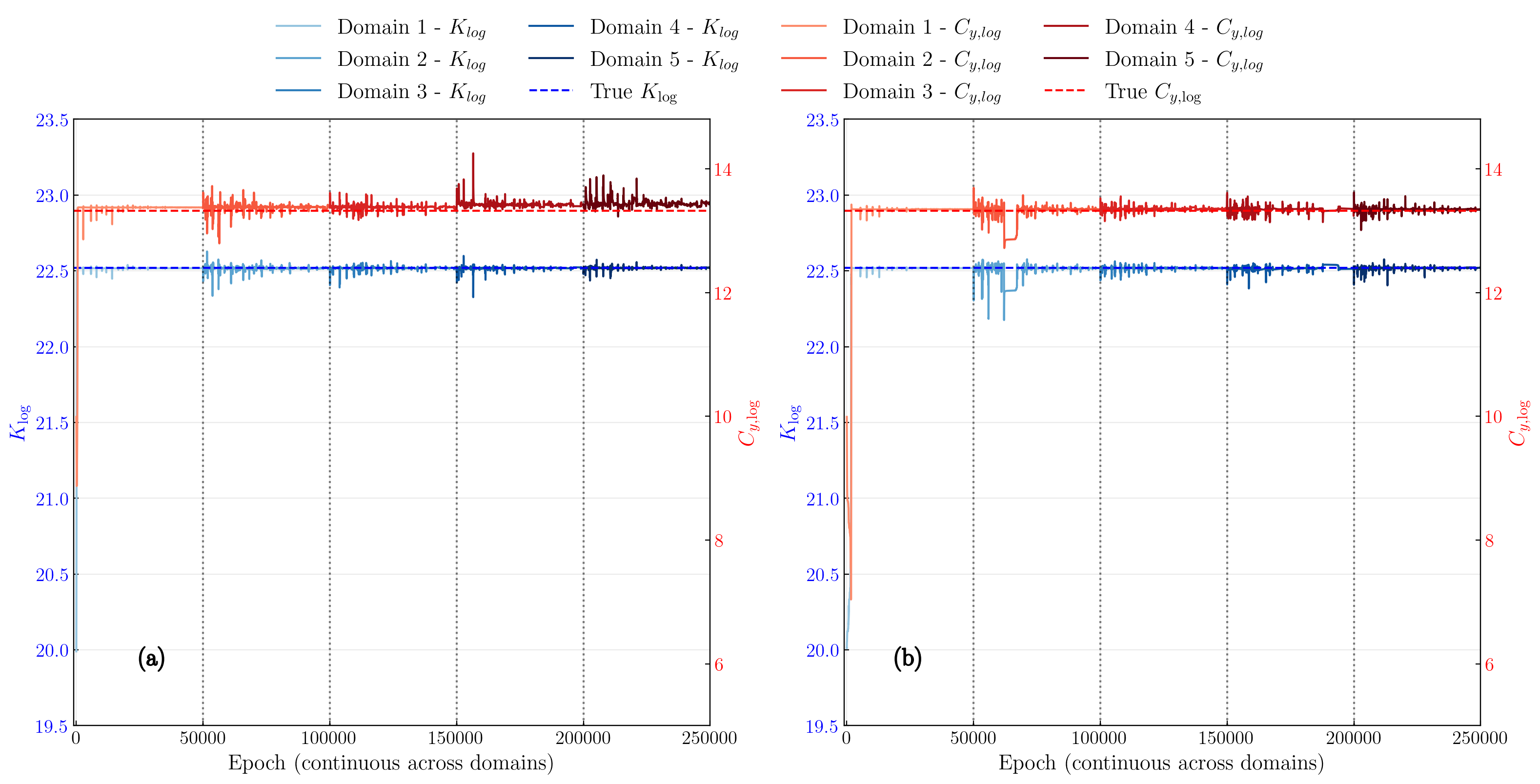}
    \caption{I-XPINN results in estimating the values of $K_{\log}$ and $C_{y,\log}$ for the (a) stable case and (b) unstable case with noisy reference data of noise intensity $\alpha_n=0.01$ over the simulation interval $\tau\in[0, 5\tau_w]$ using the history function $\Phi_H(\tau)=\sin(\tau)$. The blue and red dotted horizontal lines indicate the true values of $K_{\log}$ and $C_{y,\log}$, respectively.}
    \label{fig:xpinn-inv-1percent}
\end{figure*}

\begin{figure*}[!hbt]
    \centering
    \includegraphics[width=\linewidth]{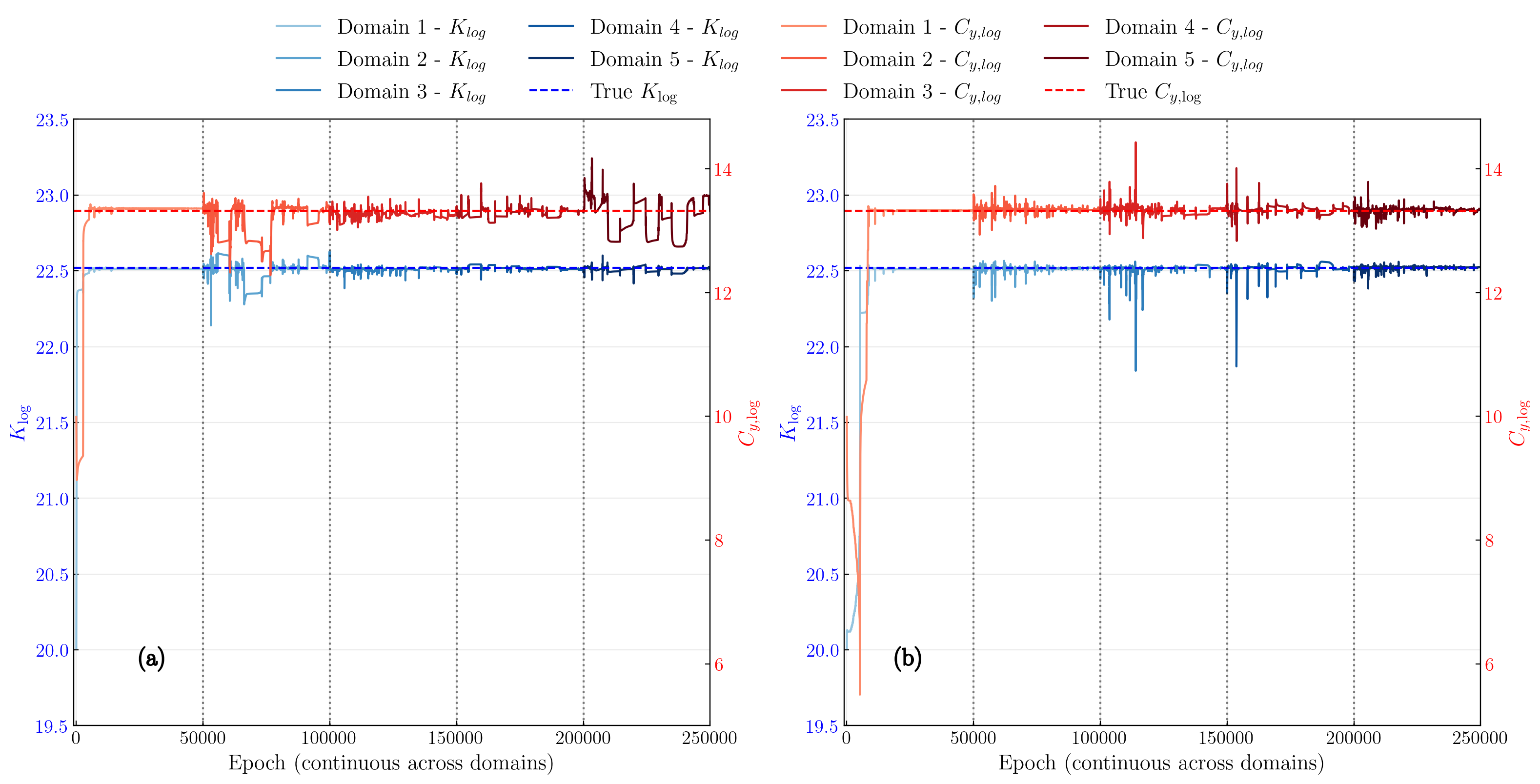}
    \caption{I-XPINN results in estimating the values of $K_{\log}$ and $C_{y,\log}$ for the (a) stable case and (b) unstable case with noisy reference data of noise intensity $\alpha_n=0.05$ over the simulation interval $\tau\in[0, 5\tau_w]$ using the history function $\Phi_H(\tau)=\sin(\tau)$. The blue and red dotted horizontal lines indicate the true values of $K_{\log}$ and $C_{y,\log}$, respectively.}
    \label{fig:xpinn-inv-5percent}
\end{figure*}

 The I-XPINN framework demonstrates the accuracy in estimating the unknown parameters and reconstructing the DDE solution over an extended time interval for both stable and unstable conditions with a non-constant history function. However, the data used in these analyses were generated from the numerical solution of the governing DDE \eqref{eq-nondim} and do not account for measurement uncertainty. In practical machining applications, the measured response is inevitably influenced by sensor noise and other sources of experimental uncertainty. Therefore, it becomes essential to evaluate whether the proposed I-XPINN framework can still achieve reliable parameter estimation and solution reconstruction when the available reference data deviate from the underlying physical response. This analysis provides a more realistic assessment of the robustness of the proposed I-XPINN method and determines whether the physics-informed formulation can distinguish the underlying dynamics from the measurement noise. To introduce measurement uncertainty, Gaussian noise is added to the reference true solution $y_{\mathrm{ref}}$ obtained from DDE23 as
\begin{equation}
    y_{\mathrm{ref}}^{\mathrm{noise}}(\tau) = y_{\mathrm{ref}}(\tau) + \sigma_n \epsilon_n(\tau),
\end{equation}
where
\begin{equation}
    \sigma_n = \alpha_n \operatorname{\mathbf{std}}(y_{\mathrm{ref}}),
\end{equation}
and $\epsilon_n \sim \mathcal{N}(0,1)$. Here, $\alpha_n$ represents the prescribed noise intensity (fraction of signal std) and $\operatorname{\mathbf{std}}$ represents the standard deviation. Importantly, the noise is introduced only in the reference data used in the data loss computation given in Eq.~\eqref{eq-data_loss}, while the governing DDE residual Eq.~\eqref{Eq:residualDDE} employed in the physics loss remains unchanged. Thus, the I-XPINN requires estimating the unknown parameters from the noisy observations while simultaneously approximating the solution to satisfy the original governing DDE. The analyses with noisy reference data are carried out, and the results are shown in Fig.~\ref{fig:xpinn-inv-1percent} and \ref{fig:xpinn-inv-5percent} respectively for $\alpha_n=0.01$ and $\alpha_n=0.05$.
\par From Fig.~\ref{fig:xpinn-inv-1percent}, it is clear that the evolution of the unknown parameters for both the stable and unstable cases, I-XPINN is able to estimate both $K_{\log}$ and $C_{y, \log}$ for the given reference data with $1\%$ noise intensity. A similar kind of trend is also seen in Fig.~\ref{fig:xpinn-inv-5percent} for the case of reference data with $5\%$ noise intensity. In this case, for both the stable and unstable cases, the evolution of $C_{y,\log}$, particularly in the first domain, starts to have values less than the given initial guess value of $10$ until the other parameter $K_{\log}$ matches closely with the true reference value. In particular, for the unstable case shown in Fig.~\ref{fig:xpinn-inv-5percent}~(b), it is evident that the value of $C_{y,\log}$ goes up to the value of $5.5$. The presence of oscillations at the beginning of each domain shown in Figs.~\ref{fig:xpinn-inv-1percent} and \ref{fig:xpinn-inv-1percent} for the noisy case is slightly higher than that of the previous case without noise fluctuations shown in Fig.~\ref{fig:xpinn-inv-both}. 

\par To evaluate and compare the performance of I-XPINN, the estimated values of $K$ and $C_y$ are compared against their true reference values using the relative percentage error defined in Eq.~\eqref{eq-percent-error}, and the corresponding results are summarised in Table~\ref{tab:error_compar_noise_iXPINN}. 
\begin{table*}[!hbt]
\centering
\begin{tabular}{c c c c c c}
\toprule
Region & $\alpha_n$ & $K_{\mathrm{pred}}\;[\times10^{9}]$ & $E_K\;\mathrm{[\%]}$ & $C_{y,\mathrm{pred}}\;[\times10^{5}]$ &$E_{C_{y}}\;\mathrm{[\%]}$ \\
\toprule
\multirow{3}{*}{Stable} & $0\%$ & $6.02017$& $0.0027 $ & $6.56324$ & $7.4179$\\
\cmidrule(lr){2-6}
& $1\%$ & $6.03042$ & $0.1731$ & $6.86907$ & $12.4235$\\
\cmidrule(lr){2-6}
& $5\%$ & $5.95266$ & $1.1186$ & $6.05781$ & $0.8541$ \\
\midrule
\multirow{3}{*}{Unstable} & $0\%$ & $6.01893$& $0.0178$ & $6.26585$ & $2.5507$\\
\cmidrule(lr){2-6}
& $1\%$ & $6.03175$ & $0.1952$ & $6.23803$ & $2.0955$\\
\cmidrule(lr){2-6}
& $5\%$ & $6.05486$ & $0.579$ & $6.18253$ & $1.1871$ \\
\bottomrule
\multicolumn{3}{r}{$K_{\mathrm{true}}=6.02\times10^{9}$,}&\multicolumn{2}{r}{$C_{y,\mathrm{true}}=6.11\times10^{5}$}&\\
\bottomrule
\end{tabular}
\caption{Comparison of relative errors of parameter estimations with I-XPINN for various noise intensities $\alpha_n$.}
\label{tab:error_compar_noise_iXPINN}
\end{table*}
The noise-free case from the previous analysis is also included in the table as $\alpha_n=0\%$ for comparison. For the noise-free data, the XPINN provides highly accurate estimates of K, with relative errors of $0.0027\%$ and $0.0178\%$ for the stable and unstable cases, respectively. The corresponding errors in $C_y$ are $7.4179\%$ and $2.5507\%$. These values are considered the baseline for analysing the robustness of the I-XPINN's parameter estimation under noisy observations. When the noise intensity $\alpha_n$ is set to $1\%$ of the standard deviation of the signal in the training data, the error in $K$ increases to $0.1731\%$ for the stable case and $0.1952$ for the unstable case. Similarly, the error in $C_y$ increases to $12.4235\%$, whereas there is a slight improvement in the unstable case, with the error decreasing from $2.5507\%$ to $2.0955\%$. This shows that measurement noise does not degrade the estimation of both parameters uniformly, instead, its impact varies with the dynamical regime and the specific parameter being identified. When the noise intensity $\alpha_n$ is increased to $5\%$, the stable case exhibits a further increase in the $K$ error $E_K$ to $1.1186\%$. Interestingly, however, the estimated $C_y=6.05781\times 10^{5}$ gives an error $E_{C_y}=0.8541\%$, which is lower than that obtained for the noise-free and $1\%$ noise cases. For the unstable case, the errors remain relatively small, with $E_K=0.579\%$ and $K_{C_y}=1.1871$. Thus, even with $5\%$ measurement noise, the I-XPINN is able to estimate the unknown parameters with $1.2\%$ error. 
\begin{figure*}[!hbt]
    \centering
    \includegraphics[width=\linewidth]{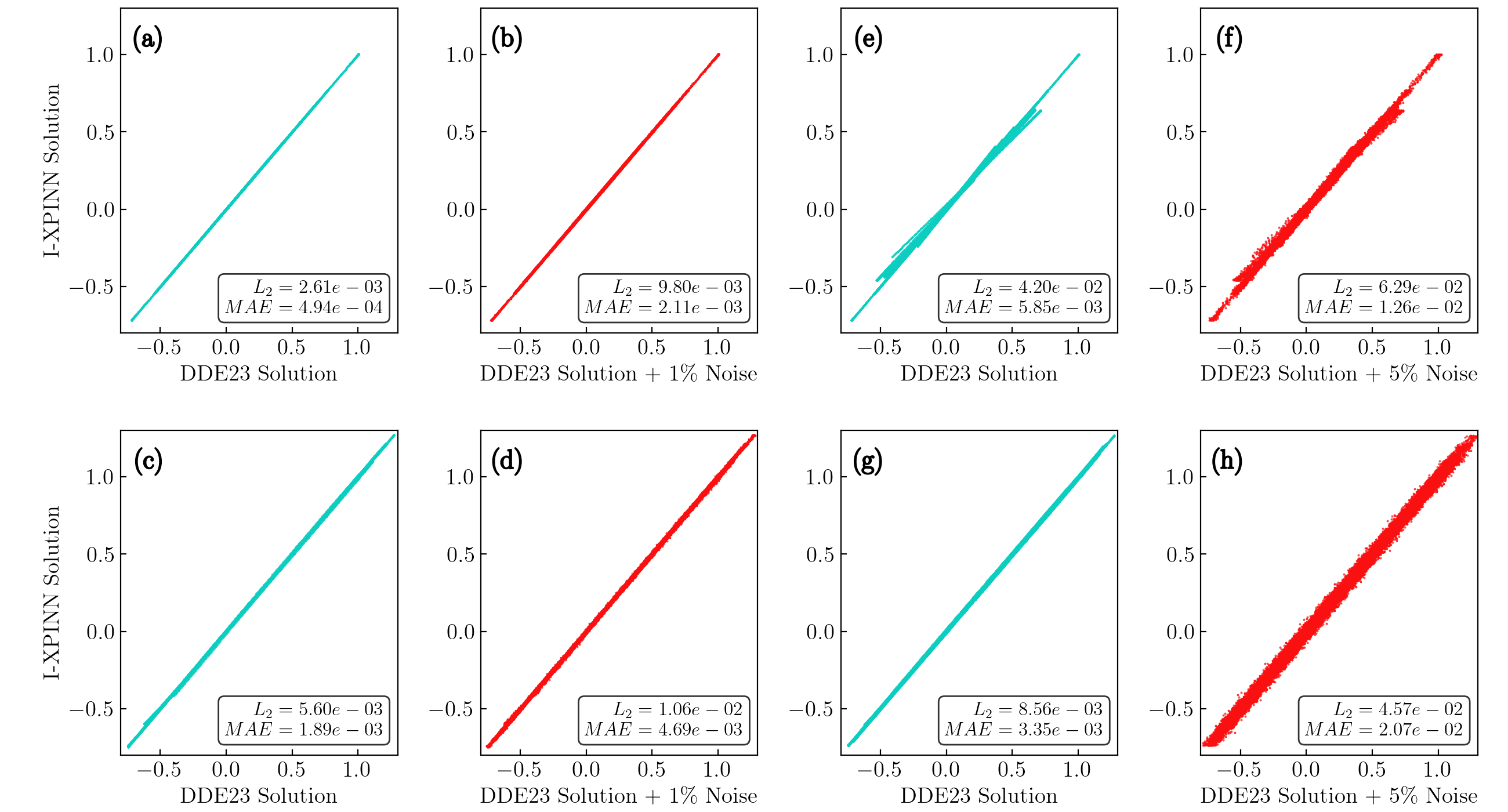}
    \caption{Scattered plots of the I-XPINN approximated solution against the reference DDE23 solutions for the stable and unstable cases. The top and bottom rows correspond to the stable and unstable cases, respectively. Figs.~(a)-(d) show the results for the I-XPINN trained with 1\% noisy reference data, while Figs.~(e)-(h) show the results for the model trained with 5\% noisy reference data. The green and red scattered points represent the noise-free DDE23 solution and the noisy reference data, respectively, and the I-XPINN-approximated solution is compared against both datasets.}
    \label{fig:xpinn-inv-scatter-noise}
\end{figure*}

\begin{table*}[!hbt]
\centering

\begin{tabular}{c c c c c c}
\toprule

    \multirow{3}{*}{
        \begin{tabular}{c}
        Error \\metric
        \end{tabular}} & \multirow{3}{*}{
        \begin{tabular}{c}
        Test data
        \end{tabular}}

&
\multicolumn{4}{c}{
    Value of $\alpha_n$ used in  $y_{\mathrm{ref}}^{\mathrm{noise}}$ during training
} \\

\cmidrule(lr){3-6}

&
&
\multicolumn{2}{c}{$1\%$}
&
\multicolumn{2}{c}{$5\%$} \\

\cmidrule(lr){3-4}
\cmidrule(lr){5-6}

&
&
Stable & Unstable & Stable & Unstable \\

\midrule

\multirow{2}{*}{$L_2$} 
& $y_{\mathrm{pred}}$ vs $y_{\mathrm{ref}}$
& $2.61\times10^{-3}$ 
& $5.60\times10^{-3}$ 
& $4.20\times10^{-2}$ 
& $8.56\times10^{-3}$ \\

& $y_{\mathrm{pred}}$ vs $y_{\mathrm{ref}}^{\mathrm{noise}}$
& $9.80\times10^{-3}$ 
& $1.06\times10^{-2}$ 
& $6.29\times10^{-2}$ 
& $4.57\times10^{-2}$ \\

\midrule

\multirow{2}{*}{MAE} 
& $y_{\mathrm{pred}}$ vs $y_{\mathrm{ref}}$
& $4.94\times10^{-4}$ 
& $1.89\times10^{-3}$ 
& $5.85\times10^{-3}$ 
& $3.35\times10^{-3}$ \\

& $y_{\mathrm{pred}}$ vs $y_{\mathrm{ref}}^{\mathrm{noise}}$
& $2.11\times10^{-3}$ 
& $4.69\times10^{-3}$ 
& $1.26\times10^{-2}$ 
& $2.07\times10^{-2}$ \\

\bottomrule
\end{tabular}
\caption{Relative $L_2$ and $MAE$ values of the I-XPINN reconstructed solution evaluated against both the noise-free DDE23 solution and the measurement noise-added reference solution.}
\label{tab:noise_solution_errors}
\end{table*}
\par The results from this analysis do not exhibit a strictly monotonic relationship between noise intensity and parameter estimation error. While $E_K$ generally increases with increasing noise, the error associated with $C_y$ shows non-monotonic behaviour. This difference may arise from the specific realisations of the random measurement noise and their interaction with the training process and the given physics-informed constraints. Therefore, the lower error values in estimating the process damping coefficient $C_y$ at higher noise intensities should not be considered evidence that noise improves the parameter-identifying capability of the I-XPINN framework. Rather, the overall results demonstrate that the XPINN remains capable of estimating the governing parameters with relatively low error, even in noisy settings. 

\par Next, the performance of the I-XPINN framework in reconstructing the DDE solution from noisy training data is evaluated, with the corresponding results presented in Fig.~\ref{fig:xpinn-inv-scatter-noise} and Table~\ref{tab:noise_solution_errors}. The predicted solution is assessed against both the noisy reference data used during training and the underlying noise-free DDE23 solution. This comparison is important because a physics-informed model should ideally avoid simply reproducing measurement noise and instead recover the physically meaningful underlying response. In Fig.~\ref{fig:xpinn-inv-scatter-noise}, the green and red scattered points represent the noise-free and noisy reference solutions, respectively, while the I-XPINN predictions are used to assess the reconstruction accuracy relative to both reference datasets.  From Fig.~\ref{fig:xpinn-inv-scatter-noise}, one can observe that the points are more scattered for the cases when the $1\%$ and $5\%$ noisy reference data are compared with the I-XPINN approximated solution than in the cases with the noise-free data. 

\par The quantitative reconstruction errors, evaluated using the relative $L_2$ error and MAE with respect to both reference datasets, are summarised in Table~\ref{tab:noise_solution_errors}. For the $1\%$ noise intensity, in the stable case, I-XPINN trained on $ 1\%$-noisy reference data achieves a relative $L_2$ error of $2.61\times10^{-3}$ and an $MAE$ of $4.94\times10^{-4}$ when evaluated on noise-free data. The error values increase to $9.8\times10^{-3}$ and $2.11\times 10{-3}$, respectively, when the prediction is directly compared with the $1\%$ noisy data. A similar trend is observed in the unstable case, where the corresponding errors against the noise-free solutions are $L_2=5.6\times 10^{3}$ and $MAE=1.89\times10^{-3}$, increasing to $L_2=1.06\times 10^{2}$ and $MAE=4.69\times10^{-3}$ when evaluated against the noisy data. At the higher $5\%$ noise intensity, the influence of measurement noise becomes more evident. When the approximated solution of the I-XPINN framework is trained on data with $5\%$ measurement noise and compared with the noise-free solutions for the stable case, the error values are $L_2=4.2\times 10^{-2}$ and $MAE=5.8465\times 10^{-3}$. When compared with the noisy observations, these errors increase further to $L_2=6.29\times10^{-2}$ and $MAE=1.26\times 10^{-2}$. For the unstable case, however, the prediction remains close to the noise-free solution, with an $L_2$ of $8.56\times 10^{-3}$ and an $MAE$ of $3.35\times 10^{-3}$. The corresponding errors against the noisy observations become $4.57\times 10^{-2}$ and $2.07\times 10^{-2}$ for $L_2$ and $MAE$ respectively.

\par A notable observation is that the errors computed against the noise-free solutions are consistently lower than those calculated against the noisy observations. This reveals that the I-XPINN does not simply attempt to reproduce the noise contained in the measurements. Instead, the physics-informed constraints enable the I-XPINN to recover the solution that more closely reflects the underlying machining dynamics. Overall, the results indicate that the I-XPINN is robust to measurement noise, although increasing the intensity of noise by a large margin would indeed lead to noticeable degradation in solution reconstruction, particularly for the stable case. The demonstrated ability to infer coupled, physically meaningful parameters directly from noisy responses establishes a foundation for extending the approach to experimentally measured machining data, as well as more complex machining configurations.

\section{Conclusion}\label{sec6}
\par This work developed physics-informed neural network (PINN) inverse identification approaches for unknown physical parameters in machining dynamics, addressing both regenerative delay and nonsmooth frictional effects. The coupled parameters of cutting force coefficient $K$ and process damping coefficient $C_y$ were simultaneously estimated using two complementary PINN frameworks. The forward problem, approximating the DDE solution using the provided physical information, was examined first. The results showed that the single-domain PINN (SPINN) formulation can approximate delayed dynamics over a relatively short time horizon, but it exhibits limitations when a non-constant history function is considered and the solution is extended over longer time intervals. Motivated by these limitations, a multi-domain eXtended PINN (XPINN) framework was introduced to improve the representation of the delayed machine dynamics over an extended temporal domain and a non-constant history function. The temporal domain was decomposed into multiple successive subdomains, with an individual neural network assigned to each subdomain. The subdomains were trained sequentially, with each completed interval’s parameters frozen before training the next, reducing the optimisation burden of learning the full delayed response and enabling the solution to be built progressively across the extended temporal domain. The forward data-driven solutions of these two PINN frameworks for the considered DDE machining system were evaluated using scatter plots and by computing the relative $L_2$ error and the mean absolute error. The proposed frameworks were evaluated for both stable (chatter-free) and unstable (chatter) machining regimes, and it was demonstrated that XPINN substantially improves solution accuracy compared with SPINN, reducing the relative $L_2$ error by approximately $73.2\%$ in the stable regime and $92.1\%$ in the unstable regime, relative to the best-performing SPINN configuration. These findings highlight the effectiveness of the XPINN formulation in improving the accuracy of the forward solution of the considered machining model.

\par Later, the loss functions and trainable parameters of these two SPINN and XPINN frameworks were modified for unknown parameter identification. The Inverse Single-domain Physics-Informed Neural Network (I-SPINN) and the Inverse eXtended Physics-Informed Neural Network (XPINN), were used to estimate unknown parameters. The results of inverse problems showed that the I‑XPINN delivers substantially improved inverse‑estimation accuracy, reducing the error in $C_y$ by about $71.5\%$ in the stable case and $95.2\%$ in the unstable case, and reducing the error in $K$ by approximately 99.6\% and 98.3\% in the respective cases, demonstrating a clear advantage over the I‑SPINN formulation. These inverse frameworks were also examined for DDE solution reconstruction using the estimated parameters, and the results showed that the I‑XPINN achieved relative $L_2$ errors on the order of $10^{-3}$, indicating its ability to maintain high prediction accuracy across both stable and unstable regimes over the extended temporal domain.

\par The robustness of the I‑XPINN framework was also assessed using noisy data generated by adding Gaussian noise to account for measurement uncertainty. The overall results show that I‑XPINN consistently maintains low parameter‑estimation error even under significant measurement noise, demonstrating strong resilience to measurement uncertainty. The robustness of the proposed I-XPINN framework with noisy data was further investigated by reconstructing the DDE solution from the I-XPINN model trained on noisy training data. The outcomes showed that the $L_2$ and mean absolute errors computed against the deterministic solution remain consistently lower than those computed against the noisy observations, indicating that the physics‑informed constraints guide the network in recovering the underlying machining dynamics of the DDE under consideration.

Beyond machining dynamics, this work provides the first systematic strategy for addressing inverse problems in nonlinear time-delay systems over extended time horizons exceeding one characteristic delay, $\tau$. The framework addresses substantial identification challenges due to the combined effects of delayed interactions, nonlinear frictional dynamics, and limited parameter observability due to measurement noise, which can severely constrain conventional parameter-estimation approaches. 

\section*{Acknowledgments}
The authors acknowledge the support from the Sonic Research IT HPC Cluster at UCD, funding from UCD Chang'an Dublin International College of Transportation, Research Ireland $22/FFP/P11457$ HarMonI, and Research Ireland funded I-Form Advanced Manufacturing Research Centre $1/RC/10295\_P2$. Vikram Pakrashi would like to acknowledge Research Ireland Rinn Energy 23/RC/12218), and Horizon Europe MSCA DN DIAMOND (101227338).

\section*{Data Availability Statement}
The data that support the findings of this study are available within the article.

\bibliographystyle{elsarticle-num-names} 

\bibliography{references.bib}
\end{document}